\documentclass[10pt,twocolumn,letterpaper]{article}
\usepackage[pagenumbers]{cvpr} 

\usepackage{footmisc}

\usepackage{adjustbox}
\usepackage{amsmath}
\usepackage{amssymb}
\usepackage{arydshln}
\usepackage{booktabs}
\usepackage{csvsimple}
\usepackage{etoolbox}
\usepackage{graphicx}
\usepackage{lipsum}
\usepackage{makecell}
\usepackage{makecell}
\usepackage{multirow}
\usepackage{pgf-pie}
\usepackage{pgffor}
\usepackage{pgfplotstable}
\usepackage{pifont}
\usepackage{tikz}
\usepackage{xspace}
\usepackage{calc}

\pgfplotsset{compat=1.7} 

\newcommand{\xmark}{\text{\ding{55}}}

\definecolor{themelight}{RGB}{193,218,239}
\definecolor{themedark}{RGB}{83,123,156}

\makeatletter
\renewcommand{\paragraph}{%
    \@startsection{paragraph}{4}%
    {\z@}{-0.5em}{-0.5em}%
    {\normalfont\normalsize\bfseries}%
}
\makeatother

\newcommand{\ucothreed}{uCO3D\xspace}

\newif\ifprintcomments
\printcommentstrue

\makeatletter
\DeclareRobustCommand\onedot{\futurelet\@let@token\@onedot}
\def\@onedot{\ifx\@let@token.\else.\null\fi\xspace}

\def\eg{\emph{e.g}\onedot}

\makeatother

\newlength\myheightinline
\newlength\mydepthinline
\settototalheight\myheightinline{Xygp}
\newcommand*\inlinegraphics[1]{%
  \settototalheight\myheightinline{Xygp}%
  \settodepth\mydepthinline{Xygp}%
  \raisebox{-\mydepthinline}{\includegraphics[height=\myheightinline]{#1}}%
}

\definecolor{cvprblue}{RGB}{134,134,255}
\definecolor{ucotd}{RGB}{92,34,25}

\usepackage[pagebackref,breaklinks,colorlinks,allcolors=cvprblue]{hyperref}

\def\paperID{2038}
\def\confName{CVPR}
\def\confYear{2025}

\title{UnCommon Objects in 3D}

\newcommand{\metaffil}{$^\infty$}

\author{
\vspace{0.12cm}
Xingchen Liu\metaffil\quad
Piyush Tayal\metaffil\quad
Jianyuan Wang\metaffil\quad
Jesus Zarzar$^+$\metaffil\quad
Tom Monnier\metaffil\\
\vspace{0.12cm}
Konstantinos Tertikas\metaffil*\quad
Jiali Duan\metaffil\quad
Antoine Toisoul\metaffil\quad
Jason Y. Zhang\textsuperscript{\textdagger}\\
\vspace{0.12cm}
Natalia Neverova\metaffil\quad
Andrea Vedaldi\metaffil\quad
Roman Shapovalov\metaffil\quad
David Novotny\metaffil
\\ \\
*NKUA, Greece
~
$^+$KAUST
~
\textsuperscript{\textdagger}Carnegie Mellon University
~
\metaffil{}Meta AI
}

\begin{document}
\twocolumn[{
\maketitle
\vspace{-0.7cm}%
\begin{center}
{\Large\inlinegraphics{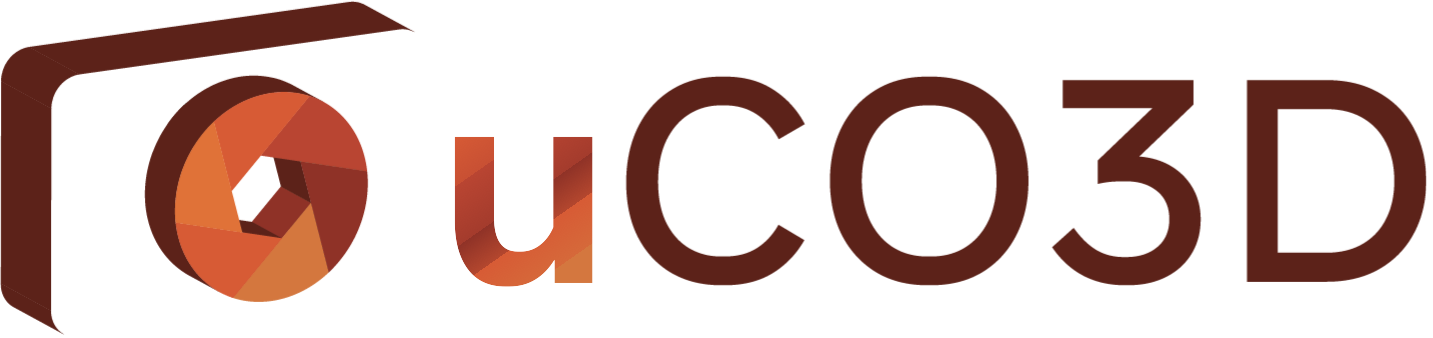}}%
\hspace{0.6cm}%
\texttt{%
{%
    \normalsize%
    \textbf{%
        \href{https://uco3d.github.io}
        {\color{ucotd}https://uco3d.github.io}%
    }%
}%
\hspace{0.6cm}%
{%
    \normalsize%
    \textbf{%
        \href{https://github.com/facebookresearch/uco3d}
        {\color{ucotd}https://github.com/facebookresearch/uco3d}
    }%
}%
}\\
\vspace{0.8cm}%
\includegraphics[width=\linewidth]{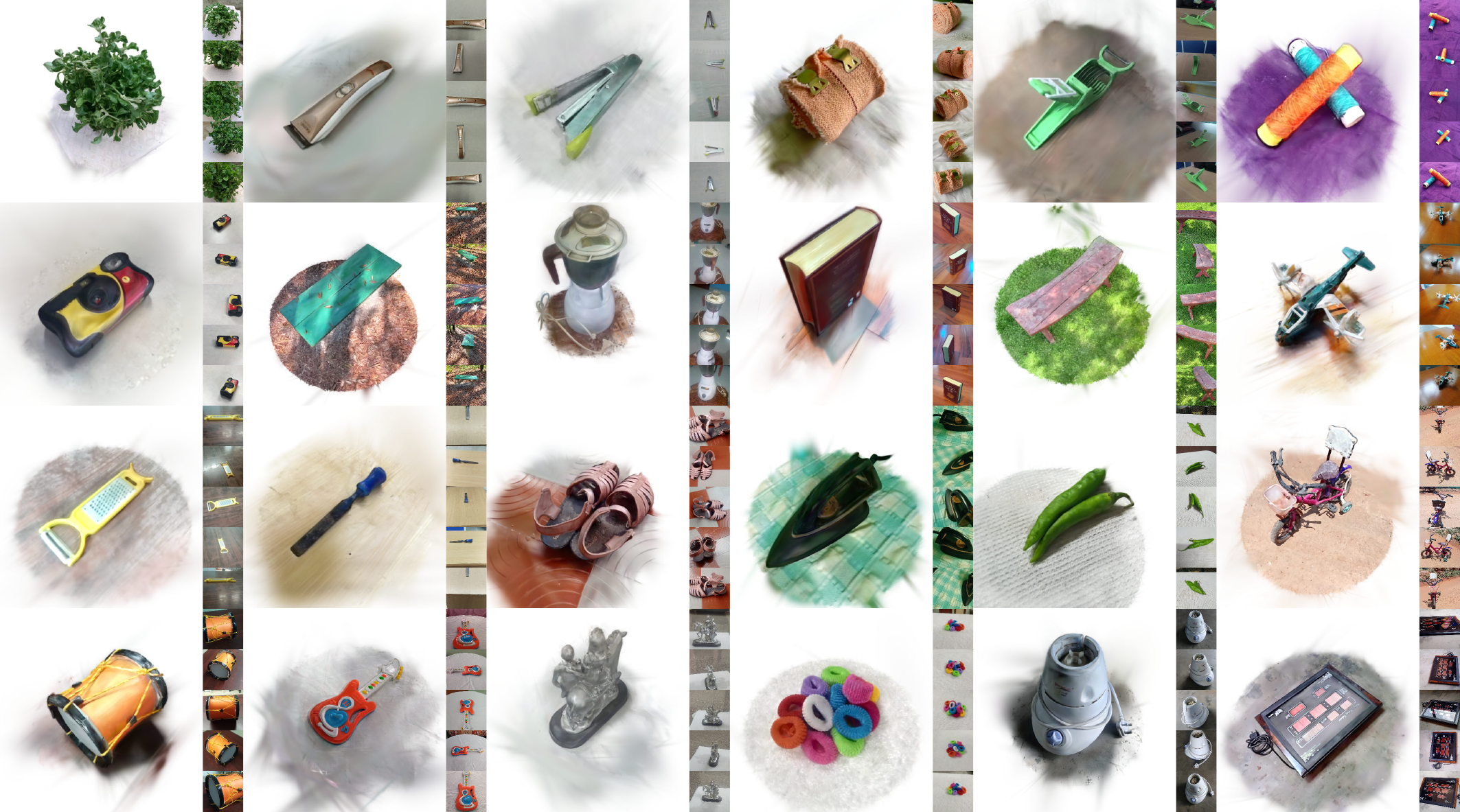}
\end{center}
\vspace{-0.5cm}
\captionsetup{type=figure}
\captionof{figure}{We introduce
\textbf{UnCommon Objects in 3D (\ucothreed)},
a large and diverse dataset of high-quality 360$^{\circ}$ videos
covering over 1,000 object categories.
Each video frame is 3D-annotated with accurate SfM cameras, point cloud,
and a 3D Gaussian Splatting reconstruction.
}\label{fig:teaser}
\vspace{1.6em}
}]
\begin{abstract}
We introduce Uncommon Objects in 3D (\ucothreed), a new object-centric dataset for 3D deep learning and 3D generative AI\@.
\ucothreed is the largest publicly-available collection of  high-resolution videos of objects with 3D annotations that ensures full-360$^{\circ}$ coverage.
\ucothreed is significantly more diverse than MVImgNet and CO3Dv2, covering more than 1,000 object categories.
It is also of higher quality, due to extensive quality checks of both the collected videos and the 3D annotations.
Similar to analogous datasets, \ucothreed contains annotations for 3D camera poses, depth maps and sparse point clouds.
In addition, each object is equipped with a caption and a 3D Gaussian Splat reconstruction.
We train several large 3D models on MVImgNet, CO3Dv2, and \ucothreed
and obtain superior results using the latter, showing that \ucothreed is better for learning applications.
\end{abstract}

\section{Introduction}%
\label{sec:intro}

\newcommand{\duster}{DUSt3r\xspace}

\begin{figure*}[ht]\centering%
\begin{minipage}[t]{0.74\textwidth}
\vspace{0pt}
\includegraphics[width=\textwidth]{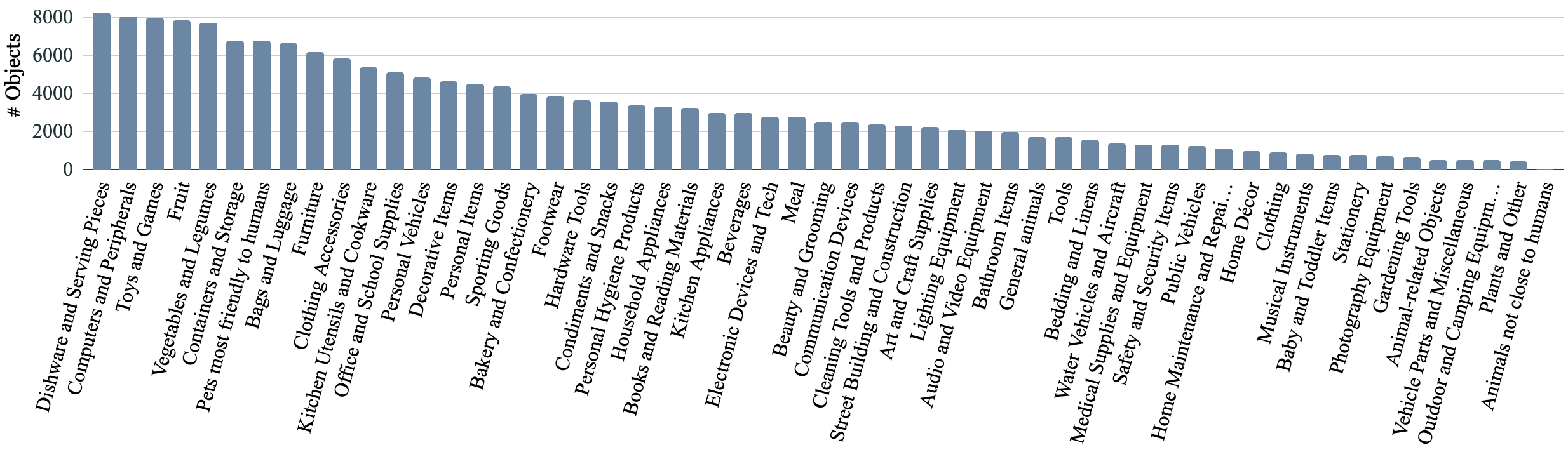}
\end{minipage}%
\hfill%
\begin{minipage}[t]{0.24\textwidth}
\vspace{0pt}
\includegraphics[width=0.99\textwidth]{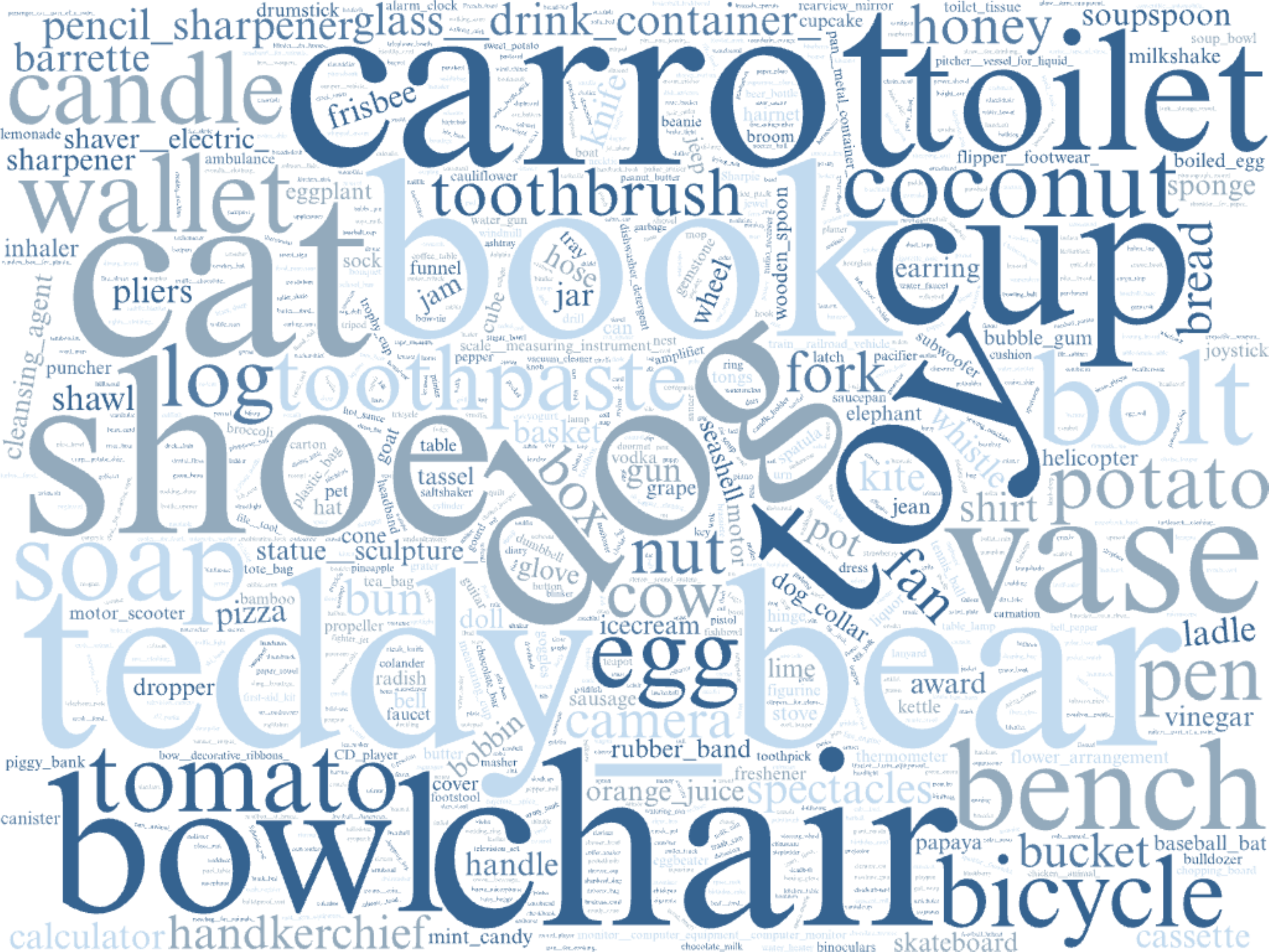}
\end{minipage}
\caption{\textbf{Statistics of \ucothreed.}
(Left) We plot the number of objects per super-category. 
In total, the dataset contains 50 super-categories, each gathering around 20 sub-categories. 
(Right) We show a word cloud of all 1,070 visual categories represented in the dataset.
}%
\label{fig:worldcloud_num_per_supercat}
\end{figure*}

The primacy of data has been the defining characteristic of the last decade of machine learning, alongside deep learning.
The most powerful models in language, speech and computer vision are simple but large deep networks trained on massive amounts of data, and then further fine-tuned on a high-quality subset of that data.
One should expect this paradigm to extend to any application of machine learning, including 3D computer vision.

However, 3D training data is much harder to come by than data for text, audio and image processing.

Seeking training data for large 3D neural networks such as the LRM of~\cite{hong24lrm:}, many have turned to synthetic datasets like Objaverse~\cite{deitke23objaverse:}.
However, synthetic data is a poor substitute for real data in applications like \emph{digital twinning}, which aims to create 3D models of real-life objects.
This is why many photorealistic reconstruction networks~\cite{wang21ibrnet:,henzler21unsupervised,szymanowicz24splatter,szymanowicz23viewset,cao2024lightplane,wu2023reconfusion} are trained using \emph{real object-centric datasets} like CO3D~\cite{reizenstein21common}, MVImgNet~\cite{yu2023mvimgnet}, GSO~\cite{downs22google} and OmniObject3D~\cite{wu2023omniobject3d}. 
Using real data is also crucial for generalization, as demonstrated by \duster~\cite{wang24dust3r:} for point map prediction and DepthAnything~\cite{yang24depth} for depth prediction, both of which are trained on numerous real datasets.
Even non-curated image datasets like the billion-scale LAION~\cite{schuhmann2022laion} are applicable to 3D vision.
For instance, text-to-3D generators~\cite{poole23dreamfusion:,shi24mvdream:,li24instant3d:,siddiqui24meta} build on large text-to-image models~\cite{blattmann23stable,dai23emu:}, which are pre-trained on such data.

Given the importance of 3D datasets, but also their relative scarcity, in this paper we ask what is the next step for real data in 3D vision.
To answer this question, we note that, while the size of a dataset is crucial, in most cases its \emph{quality} is just as important.
For example, the multi-view image generators built into current text-to-3D models~\cite{li24instant3d:,siddiqui2024meta} are notoriously sensitive to the quality of the fine-tuning data, and they are only trained on a tiny fraction of best-looking models (\eg, Instant3D~\cite{li24instant3d:} uses only about 1\% of Objaverse).
We conclude that simply contributing more low-quality  data is insufficient; instead, we need a high-quality dataset.

Based on this observation, we argue that there is a gap in the  real object-centric 3D datasets that are currently available, as none strikes the optimal balance between quality and scale.
For example, the 3D object scans in OmniObject3D~\cite{wu2023omniobject3d} and GSO~\cite{downs22google} provide very accurate geometry and textures, but only count a few thousand objects.
Conversely, datasets like CO3D~\cite{reizenstein21common} and MvImgNet~\cite{yu2023mvimgnet} contain orders-of-magnitude more objects, but lack reliable 3D scans.
Instead, they provide many \emph{views} of the objects together with lower-quality 3D cameras and point clouds reconstructed with structure-from-motion (SfM).

\begin{table}[th]
\resizebox{\columnwidth}{!}{%
\definecolor{syntheticrowcolor}{rgb}{0.9,0.9,0.9}%
\small%
\setlength{\tabcolsep}{1mm}%
\begin{tabular}{lcrrcc}
\toprule
                                                                       & Real                   & Count & \# Classes & Data type            &  Annotations  \\ \midrule
\rowcolor{syntheticrowcolor} ShapeNet~\cite{chang15shapenet}           & \color{Maroon}{\xmark} & 51k        & 55         & 3D meshes            & mesh            \\
\rowcolor{syntheticrowcolor} Objaverse~\cite{deitke23objaverse:}       & \color{Maroon}{\xmark} & 800k       & 21k        & 3D meshes            & mesh            \\
\rowcolor{syntheticrowcolor} Objaverse-XL~\cite{deitke23objaverse-xl:} & \color{Maroon}{\xmark} & 10M         & 2M        & 3D meshes            & mesh            \\
\rowcolor{syntheticrowcolor} ABO~\cite{collins2022abo}                 & \color{Maroon}{\xmark} & 8k         & 63         & 3D meshes            & mesh            \\
OmniObject3D~\cite{wu2023omniobject3d}                                 & \checkmark             & 6k         & 190        & Videos w/ meshes     & cameras, mesh   \\
GSO~\cite{downs2022google}                                             & \checkmark             & 1k         & 17         & Views w/ meshes      & cameras, mesh   \\
Objectron~\cite{ahmadyan2021objectron}                                 & \checkmark             & 15k        & 9          & Limited vp.~videos   & cameras, 3D box \\
MVImgNet~\cite{yu2023mvimgnet}                                        & \checkmark             & 220k       & 238        & Limited vp.~videos   & cameras, pcl    \\
CO3D~\cite{reizenstein21common}                                        & \checkmark             & 19k        & 50         & 360$^{\circ}$ videos & cameras, pcl    \\
CO3Dv2~\cite{reizenstein21common}                                      & \checkmark             & 40k        & 50         & 360$^{\circ}$ videos & cameras, pcl    \\
\rowcolor{themelight} \textbf{\ucothreed} (ours)                       & \checkmark             & 170k       & 1k         & 360$^{\circ}$ videos & cameras, 3DGS, caption   \\ \bottomrule
\end{tabular}
}%
\vspace{-0.5em}
\caption{%
\textbf{Overview of 3D object datasets.} We 
compare the number of objects / classes, the type of data and associated annotations.
}\label{fig:dataset_comparison}
\end{table}

In this paper, we address this gap with a new dataset, \emph{Uncommon Objects in 3D} (\ucothreed), which better balances data quality and siz (\cref{fig:dataset_comparison}).
Similar to CO3D, it comprises full-360$^\circ$ crowd-sourced videos capturing objects from all sides, annotated with cameras and point clouds using SfM.
Furthermore, \ucothreed has much greater data diversity (\cref{fig:worldcloud_num_per_supercat}) than prior alternatives as it contains objects from the 1,070 visual object categories of the LVIS~\cite{gupta19lvis:} taxonomy, which has long tails.
For reference, MVImgNet and CO3Dv2 contain only 238 and 50 categories, respectively.
These fine-grained categories are organised in super-categories, also shown in \cref{fig:worldcloud_num_per_supercat}.
Furthermore, \ucothreed contains 170k scenes, which is more than four times larger than CO3Dv2's 40k.
While this is less than MVImgNet's 220k, \ucothreed's videos cover each object from all sides, as opposed to MvImgNet's partial object captures.

Besides improving size and diversity, \ucothreed also raises the quality bar.
This was achieved by checking extensively both the collected videos and their 3D annotations.
Differently from datasets like CO3Dv2 that still contain a certain portion of low-quality videos, in \ucothreed we manually verified that each video provides full 360$^\circ$ turn-table covering all sides of the object.
Additionally, 60\%+ of the videos have 1080p+ resolution, higher than CO3Dv2.
To ensure 3D-annotation quality, we improved both the reconstruction algorithm and the reconstruction validation.
For camera reconstruction, we used VGGSfM~\cite{wang2024vggsfm}, which is currently the best SfM system available, and is more robust and accurate than COLMAP~\cite{schonberger16structure-from-motion}, used in CO3Dv2 and MvImgNet.
We also improve on CO3Dv2's active-learning camera quality evaluation by combining it with novel-view synthesis accuracy after reconstructing each scene using 3D Gaussian Splatting (3DGS)~\cite{kerbl233d-gaussian}.
The latter also guarantees that scenes are reconstructible to a high quality, which is important for training of 3D models.

We validate \ucothreed's benefits in applications.
We train two popular 3D models, LRM~\cite{hong24lrm:} and CAT3D~\cite{gao2024cat3d}, using \ucothreed and demonstrate improved results compared to training on MVImgNet and CO3Dv2, which makes \ucothreed the better data source for real object-centric 3D learning.
We also use \ucothreed to train a text-to-3D model following Instant3D's~\cite{li24instant3d:} two-stage design.
The latter requires objects to be rendered from canonical viewpoints, and thus, so far, was limited to synthetic data.
By using our 3DGS reconstructions, we `re-shoot' \ucothreed's from these viewpoints, which allows  to train a more realistic generator.
\section{Related Work}%
\label{sec:related}

\paragraph{Datasets of synthetic 3D objects.}

Historically, object-centric 3D datasets have predominantly been synthetic, composed of artist-generated 3D models.
A prominent example is ShapeNet~\cite{chang15shapenet}, with 51,000 meshes across 55 object categories.
The meshes have detailed geometries, but relatively simplistic homogeneous textures.
Other datasets such as 3D-FUTURE~\cite{fu20213d}, IKEA~\cite{lim2013parsing}, Pix3D~\cite{pix3d}, and ABO~\cite{collins2022abo} are less diverse, primarily concentrating on furniture and other consumer goods.
In contrast, ModelNet~\cite{wu153d-shapenets:}, DeepCAD~\cite{wu2021deepcad}, and ABC~\cite{koch19abc:} provide CAD models with clean geometry but lacking texture.
Objaverse~\cite{deitke23objaverse:} stands out as perhaps the most impactful dataset following ShapeNet.
It is significantly larger, comprising 800,000 artist-created 3D meshes.
Objaverse-XL~\cite{deitke23objaverse-xl:} further expands this collection to 10 million objects.
These datasets have been pivotal in advancing the development of the first 3D deep generative models, including text-to-3D~\cite{li24instant3d:,siddiqui24meta,shi24mvdream:} and image-to-3D~\cite{hong24lrm:,cao2024lightplane,xu24grm:} models. 

\paragraph{Datasets of real 3D objects.} 

Acquiring 3D data in a real-world setting presents significant challenges, resulting in a limited number of real 3D-object datasets.
Early datasets, such as Pascal3D~\cite{xiang14beyond}, contain several object categories but offer only a single view per object and only approximate 3D annotations.   
Conversely, datasets like DTU~\cite{jensen2014large}, BlendedMVS~\cite{schonberger16pixelwise}, GSO~\cite{downs22google}, OmniObject3D~\cite{wu2023omniobject3d}, Aria Digital Twin~\cite{pan2023aria}, and Digital Twin Catalog~\cite{aria_digital_twin_catalog} provide 3D scans of objects, featuring high-quality 3D geometry and textures, but containing only a few thousand objects. 
 
The use of 3D scanners significantly restricts the scale of data acquisition; consequently, other datasets capture multi-view turntable-like videos of objects using consumer cameras.
CO3D and CO3Dv2~\cite{reizenstein21common} crowd-sourced 40,000 360$^{\circ}$ object videos, providing 3D annotations by reconstructing point clouds and cameras using COLMAP SfM~\cite{schonberger16structure-from-motion}. 
MvImgNet~\cite{yu2023mvimgnet} collected even more videos (220,000) across more object categories (238), but their videos capture objects only partially, preventing full reconstruction. 
Objectron~\cite{ahmadyan2021objectron} is similar to MvImgNet, but with fewer videos (10,000).
A common challenge is that large-scale datasets often rely on SfM for video reconstruction, which can lead to imprecise 3D annotations.
\ucothreed also employs SfM, but using VGGSfM, which has greater accuracy than COLMAP, and with a more reliable data validation setup. 
Furthermore, \ucothreed is five times larger and significantly more diverse than CO3Dv2, encompassing 20 times more visual categories, and provides caption and 3D Gaussian Splat reconstructions of each object.

\begin{figure*}[ht!]\centering\small%
\includegraphics[width=\linewidth]{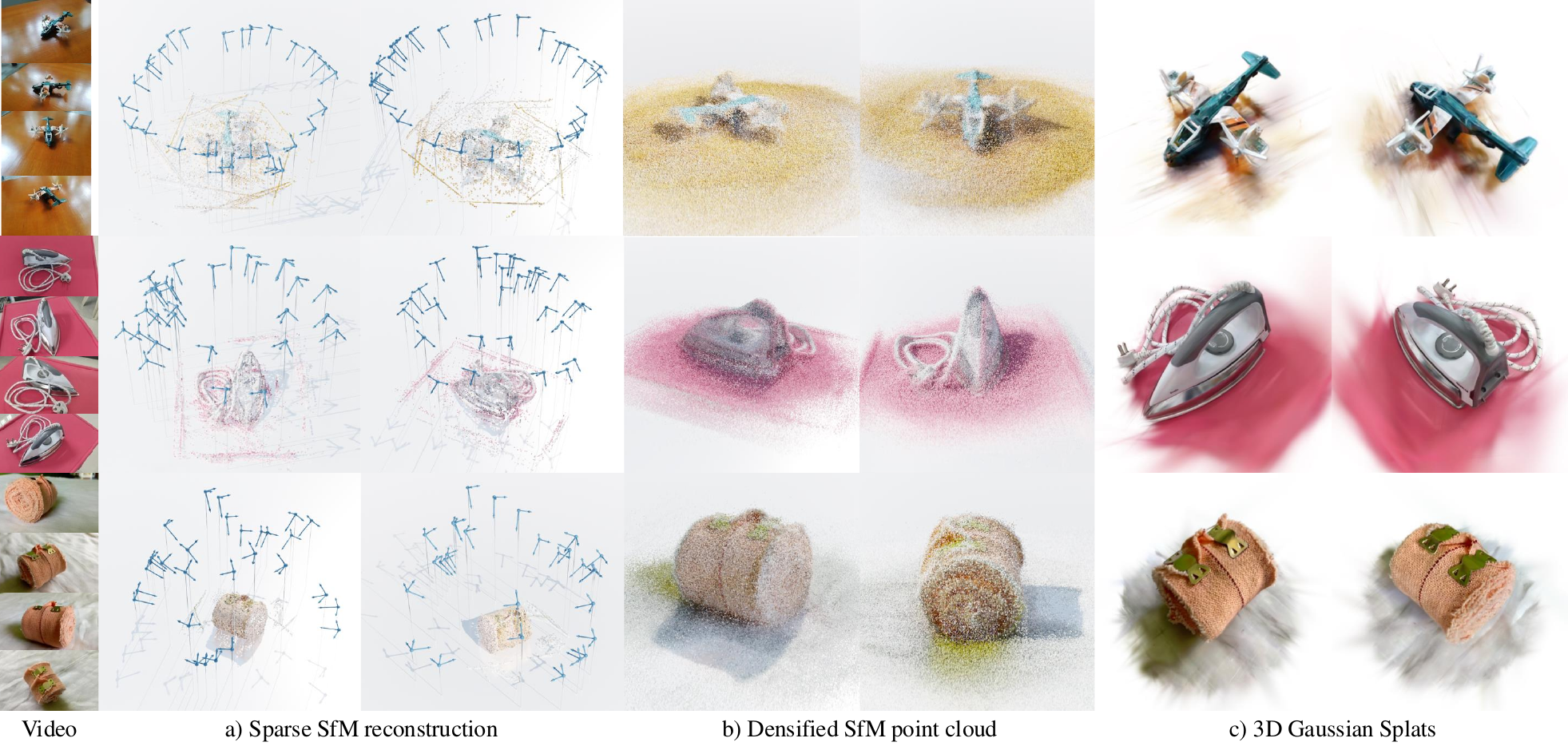}
\vspace{-0.2cm}
\caption{\textbf{Data annotation overview.} Each scene in \ucothreed is reconstructed in three different ways:
a) per-frame cameras with sparse point cloud calculated by VGGSfM~\cite{wang2024vggsfm},
b) semi-dense point cloud comprising triangulations of additional denser tracks from VGGSfM's tracker,
c) 3D Gaussian Splat~\cite{kerbl233d-gaussian} reconstruction optimized separately for each scene.
}%
\label{fig:uco3d_reconstructions}
\end{figure*}

\paragraph{Applications.}

In order to assess the quality of \ucothreed, we measure how it benefits a number of popular downstream applications.
First, we consider feedforward few-view 3D reconstruction models.
Among those, LRM~\cite{hong24lrm:} is a transformer that maps an input image to a neural radiance field supported by a triplane~\cite{chan22efficient}.
LightplaneLRM~\cite{cao2024lightplane} adds splatting layers and a memory-efficient renderer.
Further extensions use different representations like 3D Gaussian Splats~\cite{zou23triplane, xu24grm:,tang24lgm:,zhang24gs-lrm:} and meshes~\cite{xu24instantmesh:, wei24meshlrm:}.

We also consider text-to-3D generators, which create 3D assets from text, and focus on the two-stages approach of Instant3D~\cite{li24instant3d:}.
This is based on training a text-to-multi-view diffusion model~\cite{rombach22high-resolution,dai23emu:} which generates several 2D views of the object, followed by a 3D reconstruction network that outputs the 3D asset, all in a matter of seconds.
The multi-view diffusion is improved in ViewDiff~\cite{hollein2024viewdiff}, MVDiffusion~\cite{tang24mvdiffusion:}, IM-3D~\cite{melas-kyriazi24im-3d}, CAT3D~\cite{gao2024cat3d} and many others.
AssetGen~\cite{siddiqui2024meta} further extends Instant3D by modelling material properties instead of baking in the radiance function and adds a texture refiner that outputs relightable PBR textures.
As an illustration, we use \ucothreed to train a model like CAT3D, which results in better new-view synthesis than the one trained on alternative datasets.
We also show that the Gaussian Splat reconstructions provided with \ucothreed can supervise, for the first time, an Instant3D-like pipeline using solely real-life data.
\section{Uncommon Objects in 3D}%
\label{sec:uco3d}

In this section, we introduce \ucothreed, our new dataset of real-life 3D objects.
\ucothreed comprises 360$^{\circ}$ turn-table-like videos of objects, crowdsourced and annotated with 3D cameras, point clouds, 3D Gaussians, and textual captions.

Compared to older datasets like CO3Dv2~\cite{reizenstein21common}, \ucothreed comes with many improvements.
First, \ucothreed is much larger and more diverse than CO3Dv2:
it contains more than 1k different categories and more than 170k objects, compared to the 50 and 38k of CO3Dv2.
While CO3Dv2's categories are taken from MS COCO~\cite{lin14microsoft}, the categories in \ucothreed are taken from the LVIS~\cite{gupta2019lvis} taxonomy.
Hence, we inherit the LVIS focus on covering the long-tail of the visual-category distribution.
To simplify data analysis, we grouped the 1k+ LVIS categories to 50 super-categories, each containing approximately 20 subcategories.
\Cref{fig:worldcloud_num_per_supercat} shows the number of videos collected per super-category, and the LVIS category distribution.

\begin{figure}\centering\small%
\includegraphics[width=1.5cm]{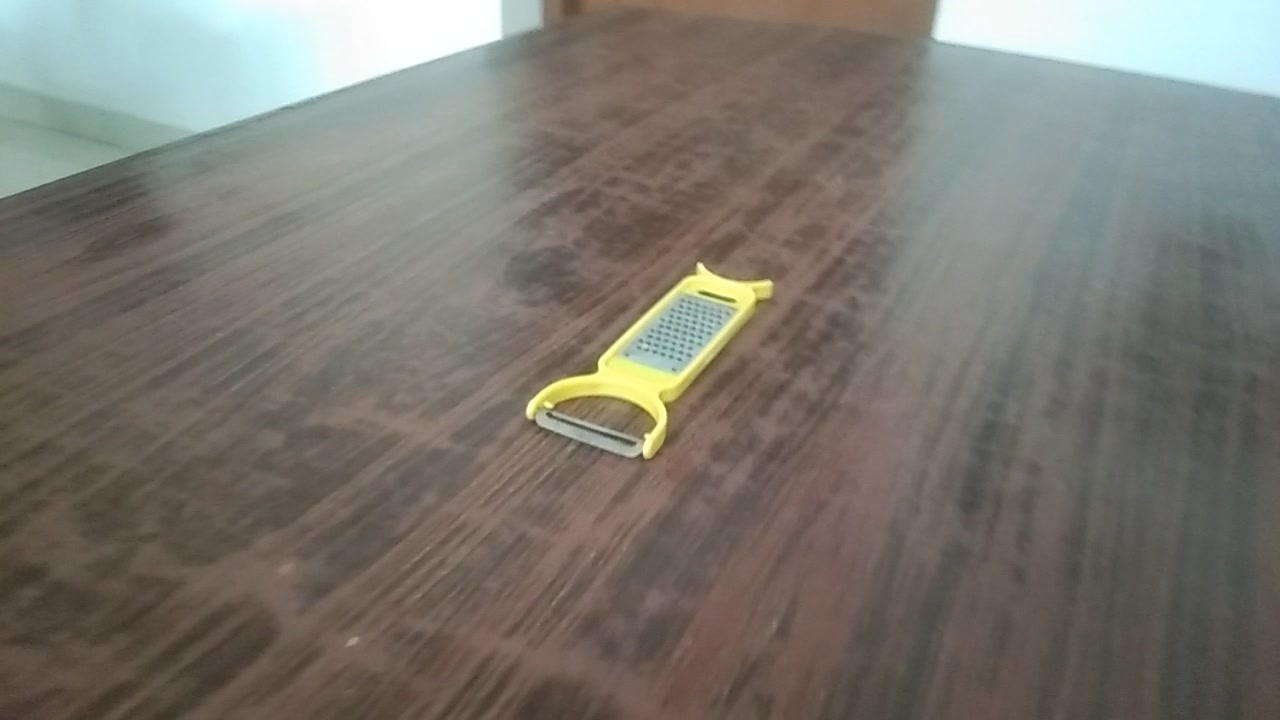}%
\includegraphics[width=1.5cm]{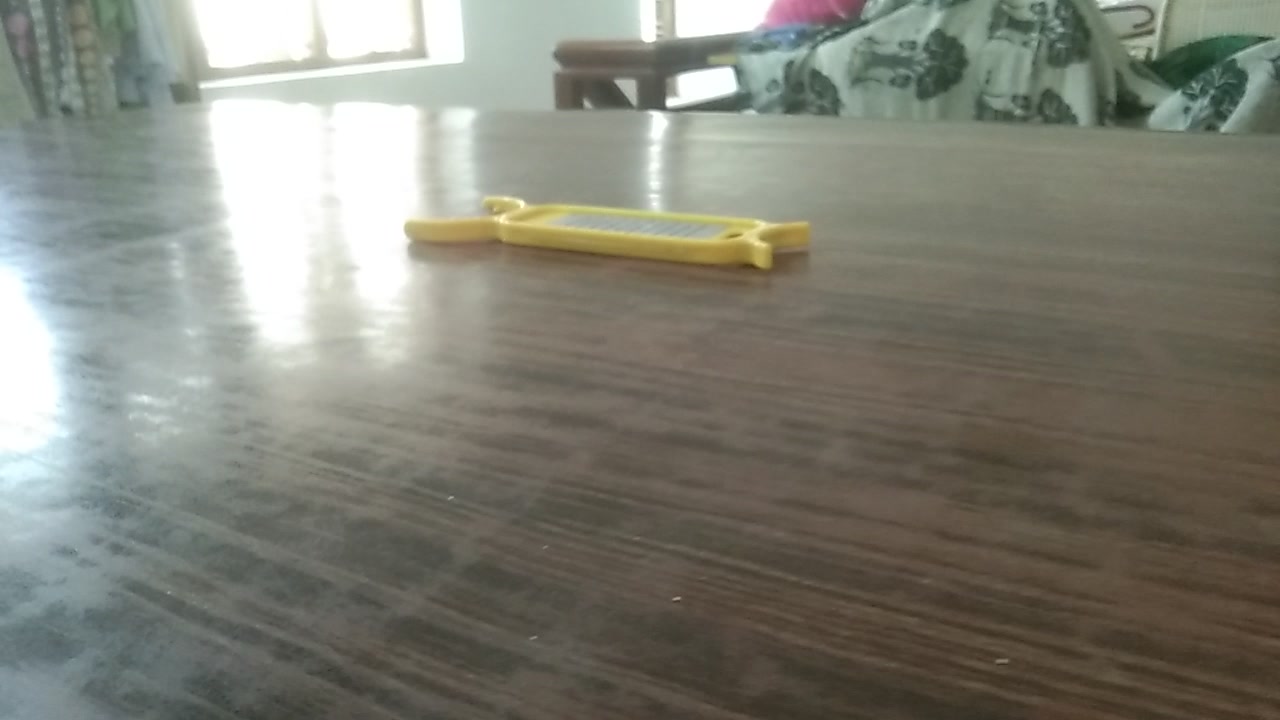}%
\includegraphics[width=1.5cm]{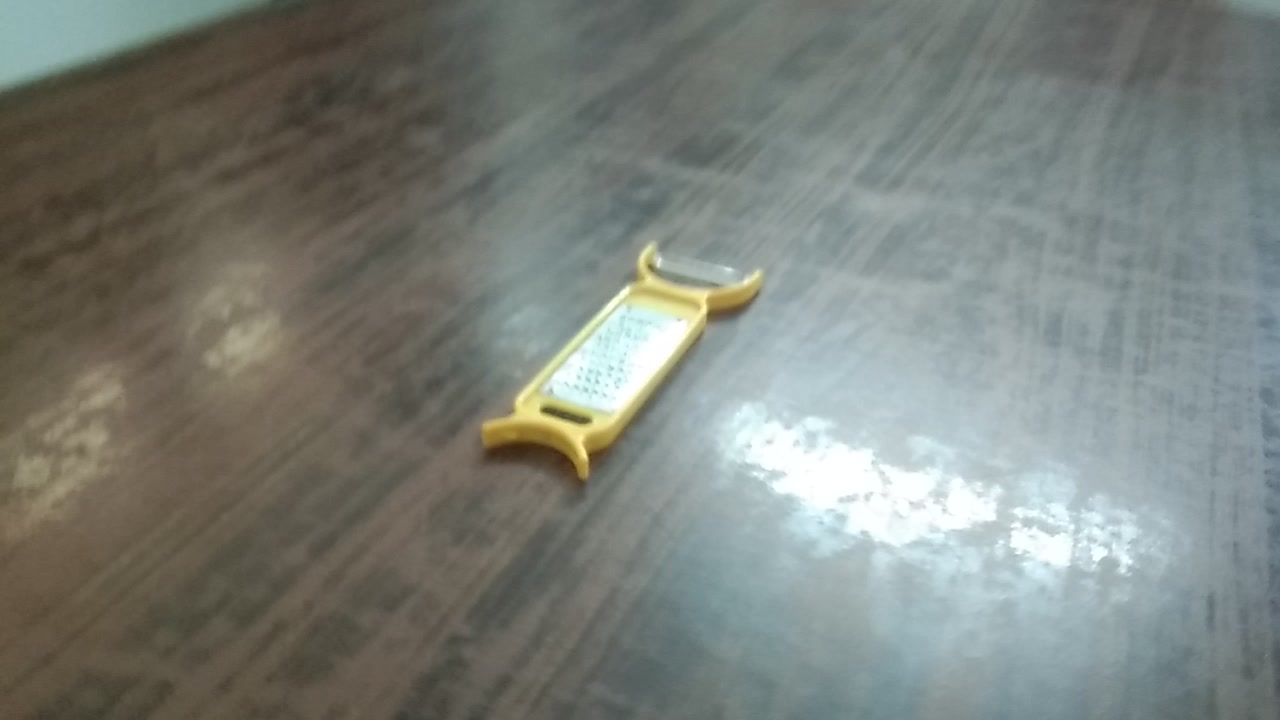}%
\includegraphics[width=1.5cm]{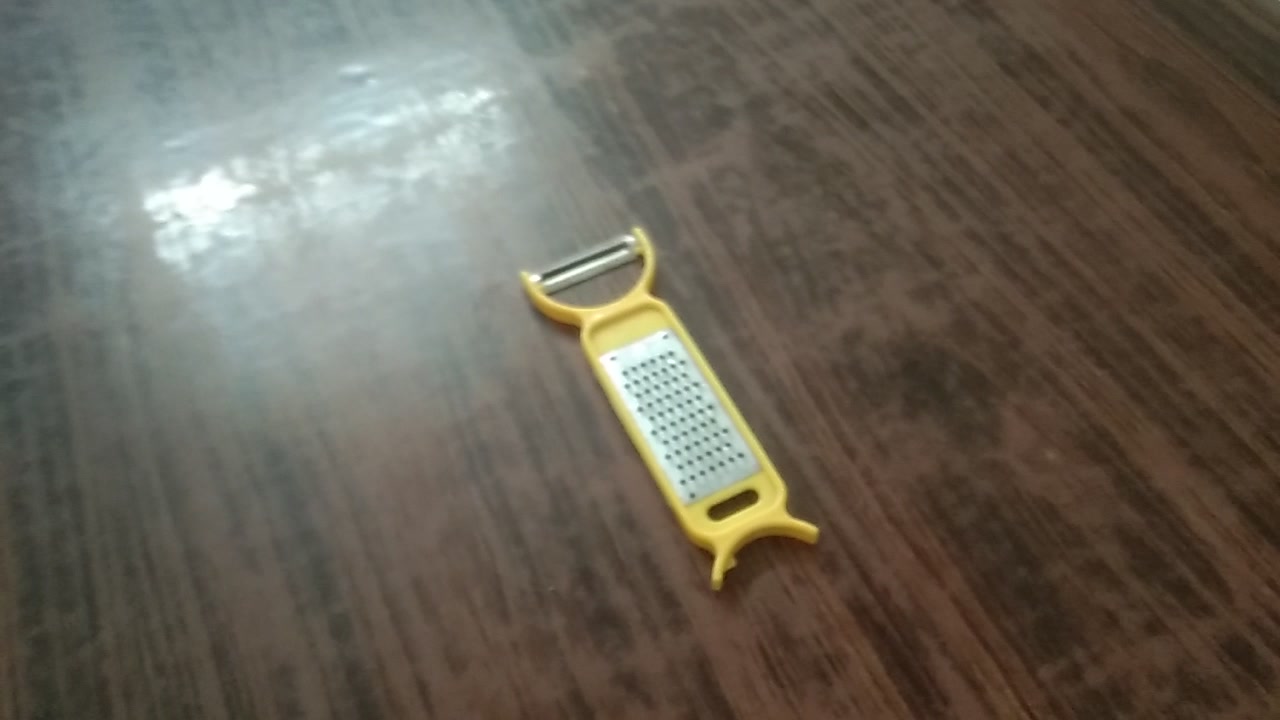}%
\includegraphics[width=1.5cm]{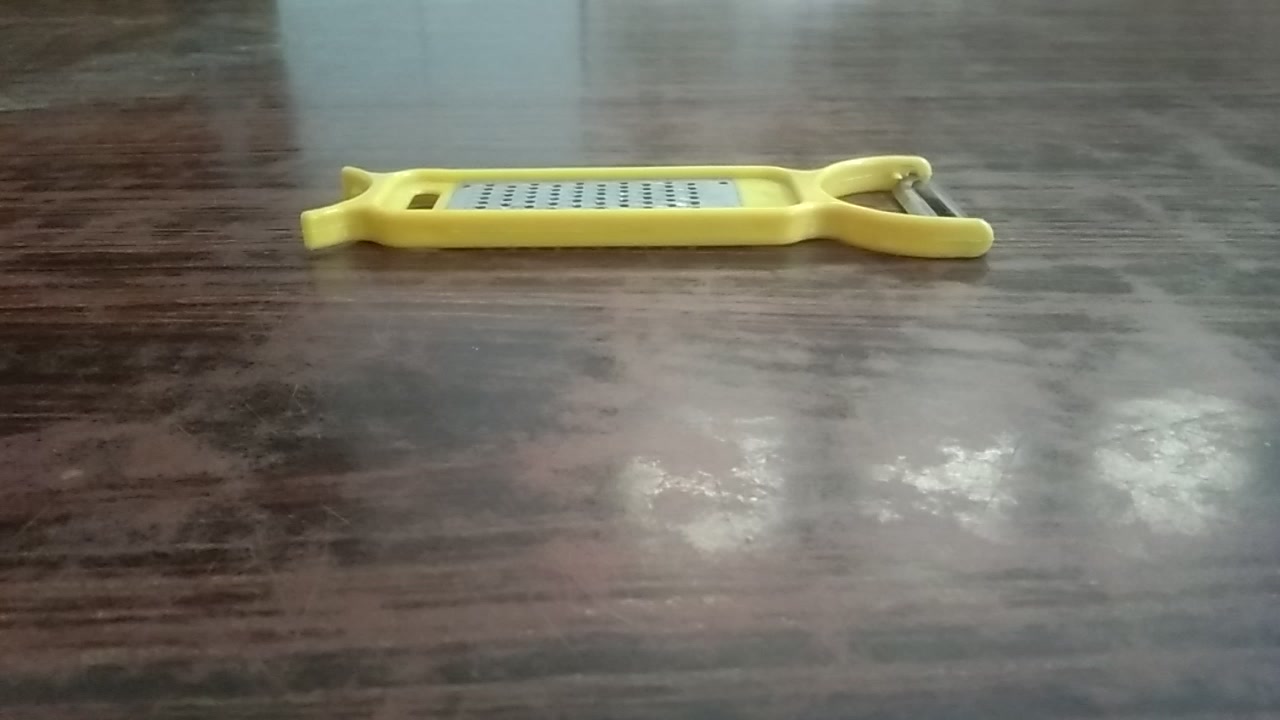}\\
\vspace{-0.1cm}a) Input video\vspace{0.1cm}\\ 
\includegraphics[height=2.3cm]{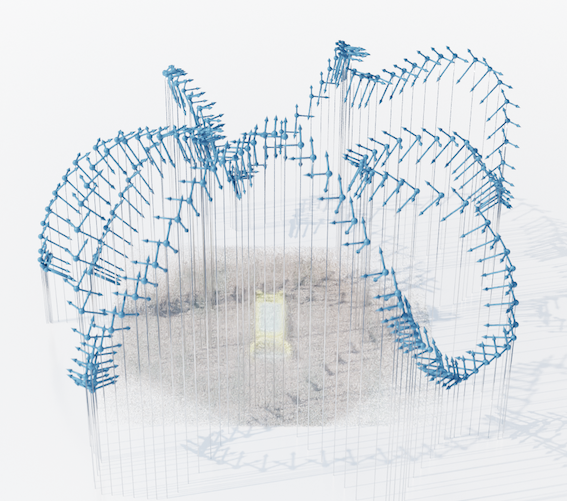}%
\includegraphics[height=2.3cm]{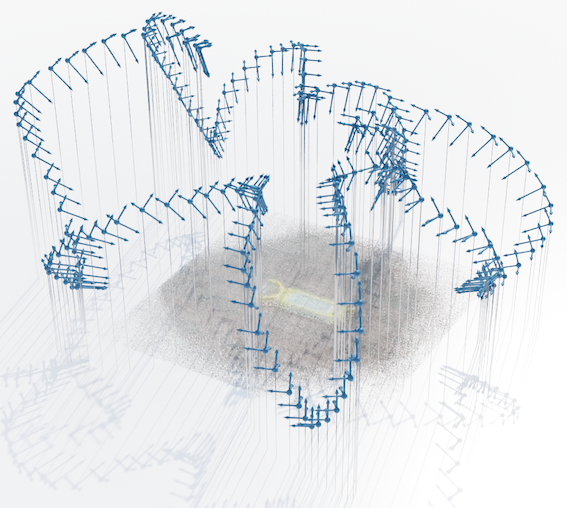}%
\includegraphics[height=2.3cm]{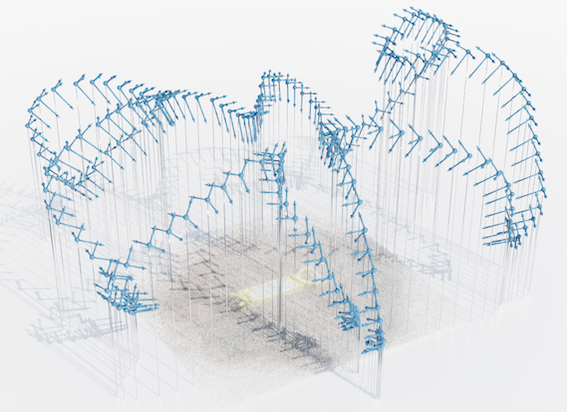}\\
\vspace{-0.1cm}b) Sine-wave camera trajectory\\ 
\vspace{-0.2cm}
\caption{\textbf{Data collection example.} For each video, the cameras follow a sine-wave trajectory to ensure good viewpoint coverage.}
\label{fig:sine_wave_cameras}
\end{figure}

Second, \ucothreed improves quality significantly compared to CO3Dv2, ensuring that videos are of high resolution, cover each object well, and that the 3D annotations are accurate.
\ucothreed also contains rich textual descriptions of each object, missing in other datasets, and useful to train large generative models.
It also comes with additional 3D Gaussian Splat reconstructions of all objects, each rigidly aligned to a canonical object-centric reference,
which make it possible to re-render the dataset from a fixed, canonical set of cameras, simulating synthetic data acqiusition, which is very useful for training generative models \cite{li24instant3d:,siddiqui2024meta}.

\paragraph{Dataset collection.}

Videos of objects were captured by workers on Amazon Mechanical Turk.
To ensure high video quality, workers were required to submit videos of a sufficient resolution.
As a result, more than 60\% of videos in \ucothreed are of 1080p resolution or higher, compared to 33\% in CO3Dv2.
Furthermore, to aid the 3D reconstruction, workers followed a sine-wave capture trajectory instead of the plain circular trajectory of CO3Dv2, ensuring varying camera elevations (c.f.~\cref{fig:sine_wave_cameras}).
Finally, each video was individually manually assessed to make sure that it adheres to these requirements, a process more rigorous than the rough eyeballing used in CO3Dv2~\cite{reizenstein21common}.

\paragraph{Video object segmentation.}

\begin{figure*}[t]
\centering
\includegraphics[width=\linewidth]{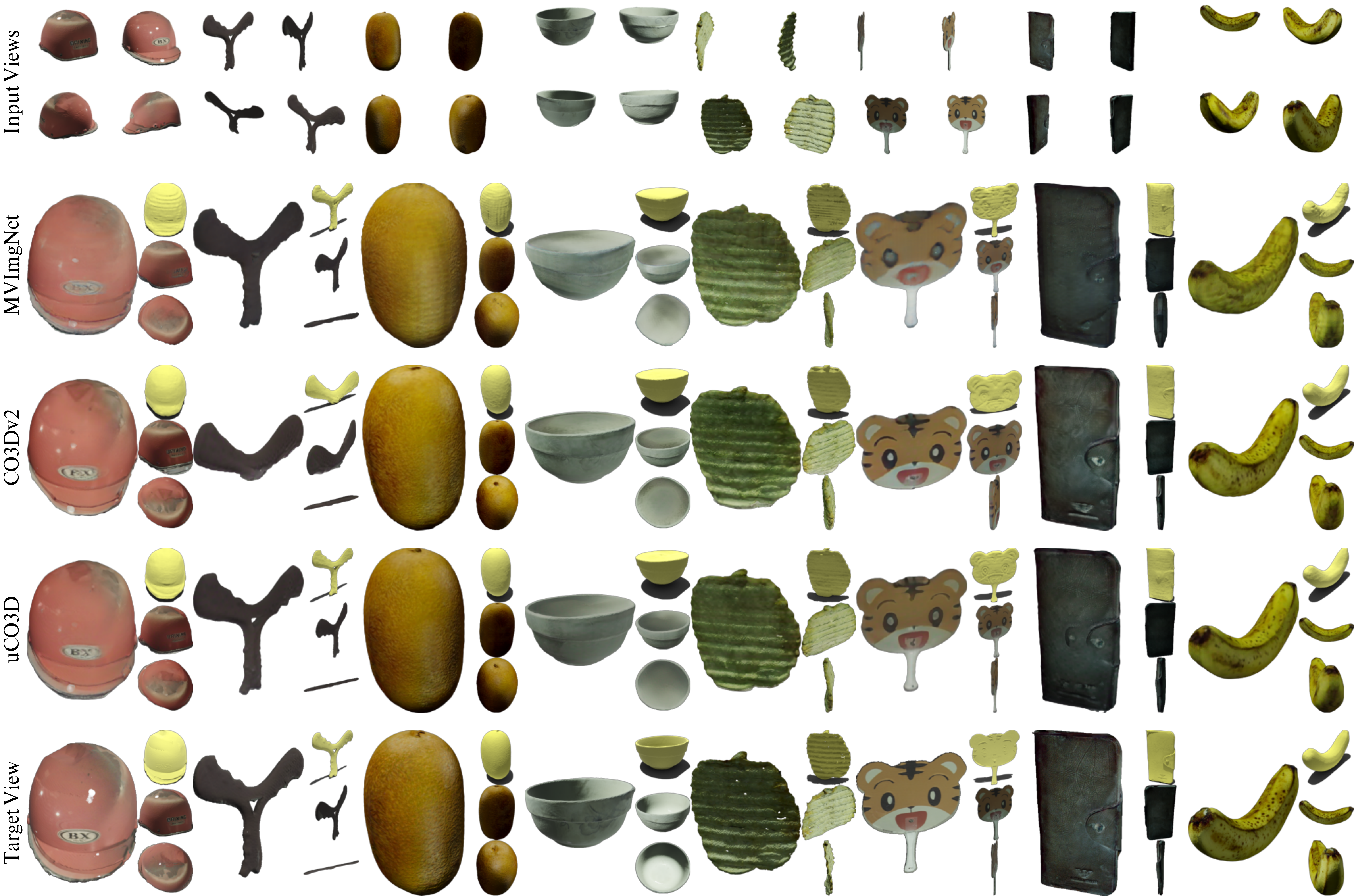}
\caption{\textbf{3D reconstruction comparison}. We show results of LightplaneLRM~\cite{cao2024lightplane} models trained on MVImgNet, CO3Dv2 and \ucothreed.}%
\label{fig:lrm}
\end{figure*}
\begin{table}[t]
\small%
\centering%
\resizebox{\linewidth}{!}{%
\setlength{\tabcolsep}{3pt}%
\begin{tabular}{@{}lrrrrrr@{}}\toprule
 & \multicolumn{3}{c}{\textbf{OmniObject3D}} & \multicolumn{3}{c}{\textbf{StanfordORB}} \\
 Train dataset & LPIPS$\downarrow$ & PSNR$\uparrow$ & IoU$\uparrow$ & LPIPS$\downarrow$ & PSNR$\uparrow$ & IoU$\uparrow$ \\ \midrule
MVImgNet~\cite{yu2023mvimgnet}                          & 0.109 & 23.39 & 0.928 & 0.070 & 24.451 & 0.939 \\
CO3Dv2~\cite{reizenstein21common}                       & 0.095 & 23.62 & 0.926 & \textbf{0.056} & 25.617 & 0.956 \\
\ucothreed (ours)                      & \textbf{0.093} & \textbf{24.61} & \textbf{0.946} & 0.057 & \textbf{25.715} & \textbf{0.957} \\
\bottomrule
\end{tabular}
}%
\caption{\textbf{3D reconstruction benchmark.} We compare LightplaneLRM~\cite{cao2024lightplane} models trained on CO3Dv2, MVImgNet,
and \ucothreed. We report novel-view synthesis performances on OmniObject3D~\cite{wu2023omniobject3d} and StanfordORB~\cite{kuang2023stanfordorb}.
\label{tab:lrm}}
\end{table}

We used text-conditioned Segment-Anything (langSAM)~\cite{kirillov23segment,guerrero:lang-sam} to segment the object of interest in each video frame given text-conditioning in form of the object-category name, which had been provided by Turkers at collection time.

To improve frame-to-frame consistency, CO3Dv2 used a simple Viterbi algorithm, which often led to segmentation flickering, impairing the final 3D reconstruction quality.
Instead, in \ucothreed, we refine the SAM segmentations with state-of-the-art deep video-segmenter based on XMem~\cite{cheng2022xmem}, leading to more temporally-stable object segmentations.

\paragraph{3D annotation with VGGSfM.}

For each video, we use the state-of-the-art VGGSfM~\cite{wang2024vggsfm} Structure from Motion (SfM) system to estimate the parameters of the cameras (intrinsic and extrinsic) for 200 frames sampled uniformly.
VGGSfM also outputs a sparse 3D point cloud, and its denser version obtained by triangulating additional 3D points from VGGSfM's tracker.
Examples of sparse and densified SfM point clouds are shown in \cref{fig:uco3d_reconstructions}.

\paragraph{Scene alignment.}

While the coordinate system of VGGSfM cameras is defined only up to a rigid transformation, it is crucial for applications like generation and reconstruction to train on a dataset of rigidly aligned objects.
We thus align all objects so they have a horizontal ground plane, similar scale, centring, and orientation.
Details of the scene alignment procedure are in the supplementary material.

\paragraph{Gaussian Splat reconstruction.}

\def\testimgrek{wc_re10k_rf_id00025}
\def\testimgllff{wc_llff_id00000}
\def\testimgdtu{wc_dtu_id00030}
\def\testimgmip{wc_mipnerf360_id00024}

\newcommand{\fourwide}{1.6in}
\newcommand{\magicwidth}{1.4in}
\newcommand{\magicthumbwidth}{0.45in}
\newcommand{\boxrek}[1]{
   \begin{tikzpicture}[every node/.style={inner sep=0,outer sep=0}]
     \node (img) at (0,0) {#1};
     \draw[color=red,thick] (-1.763,-0.985) rectangle (-0.8, -0.5);
   \end{tikzpicture}
 }
\newcommand{\boxllff}[1]{
   \begin{tikzpicture}[every node/.style={inner sep=0,outer sep=0}]
     \node (img) at (0,0) {#1};
     \draw[color=red,thick] (-0.7,0.45) rectangle (0., -0.);
   \end{tikzpicture}
 }
\newcommand{\boxdtu}[1]{
   \begin{tikzpicture}[every node/.style={inner sep=0,outer sep=0}]
     \node (img) at (0,0) {#1};
     \draw[color=red,thick] (-0.6,0.7) rectangle (0.2, -0.4);
   \end{tikzpicture}
 }
\newcommand{\boxmip}[1]{
   \begin{tikzpicture}[every node/.style={inner sep=0,outer sep=0}]
     \node (img) at (0,0) {#1};
     \draw[color=red,thick] (0.29,-0.2) rectangle (1., -1);
   \end{tikzpicture}
 }

\begin{figure*}[t]
    \centering%
    \small%
    \begin{tabular}{@{}c@{\,\,\,}c@{}c@{\,}c@{\,}c@{\,}c@{\,}c@{}} 
          & & Inputs & MVImgNet & CO3Dv2 & uCO3D & Ground-truth  \\
          \multirow{4}{*}[+4.8em]{\rotatebox{90}{
    \small$\underleftrightarrow{\hspace{20pt}\text{\small Harder}\hspace{60pt}\text{\small 
\textit{Dataset Difficulty}}\hspace{60pt}\text{\small Easier}\hspace{20pt}}$}}
         & \multirow{1}{*}[+5em]{\rotatebox{90}{\textbf{RE10K}~\cite{zhou2018stereo}}}
         &
         {
           \renewcommand{\arraystretch}{0.31}
           \begin{tabular}[b]{@{}c@{}}%
             \includegraphics[width=\magicthumbwidth]{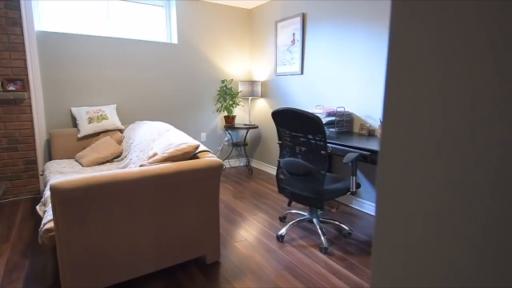} 
             \\
             \includegraphics[width=\magicthumbwidth]{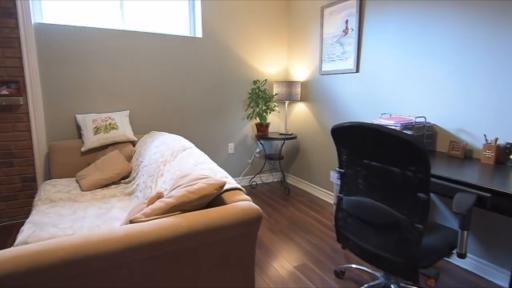} 
             \\
             \includegraphics[width=\magicthumbwidth]{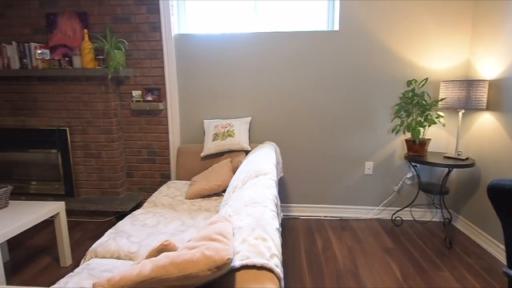}
             \\
           \end{tabular}
         }
         &
         \boxrek{
         \includegraphics[width=\magicwidth]{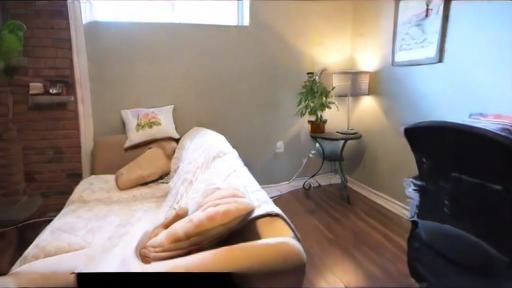} 
       }
         &
         \boxrek{
           \includegraphics[width=\magicwidth]{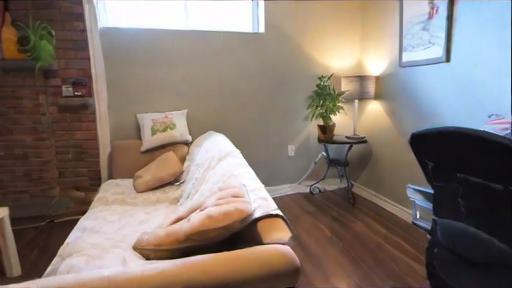} 
         }
         &
         \boxrek{
         \includegraphics[width=\magicwidth]{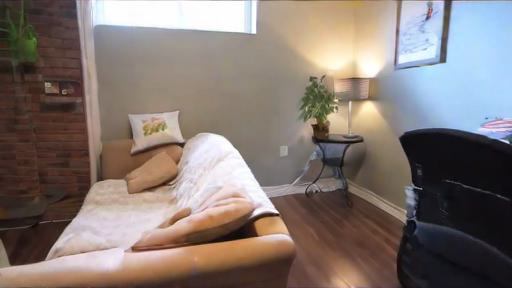} 
       }
         &
         \boxrek{
         \includegraphics[width=\magicwidth]{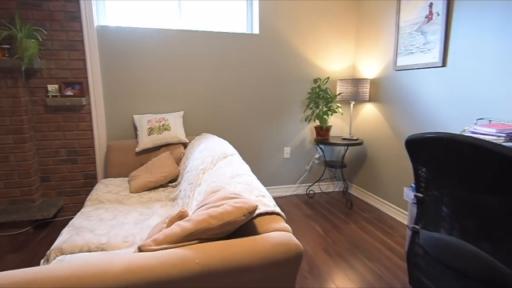}
       }
         \\

         & \multirow{1}{*}[+5.3em]{\rotatebox{90}{\textbf{LLFF}~\cite{mildenhall2019llff}}}
         &
         {
           \renewcommand{\arraystretch}{0.35}
           \begin{tabular}[b]{@{}c@{}}%
             \includegraphics[width=\magicthumbwidth]{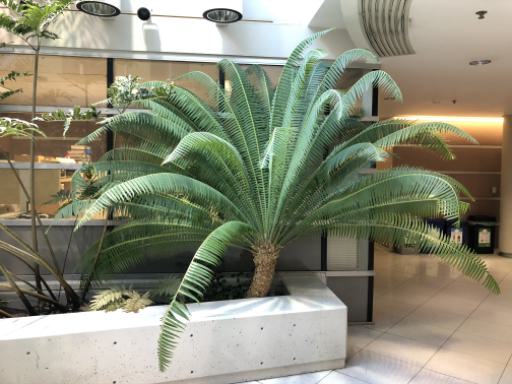} 
             \\
             \includegraphics[width=\magicthumbwidth]{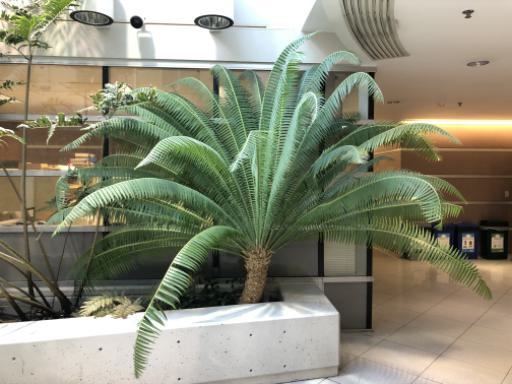} 
             \\
             \includegraphics[width=\magicthumbwidth]{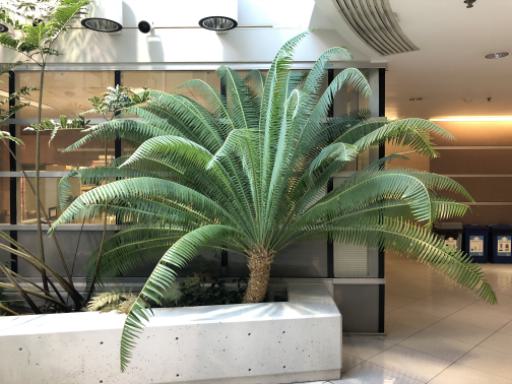}
             \\
           \end{tabular}
         }
         &
         \boxllff{
         \includegraphics[width=\magicwidth]{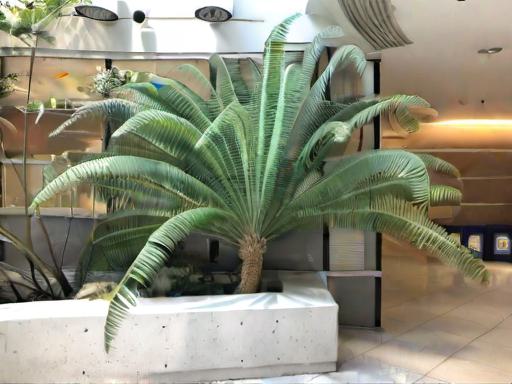} 
       }
         &
         \boxllff{
         \includegraphics[width=\magicwidth]{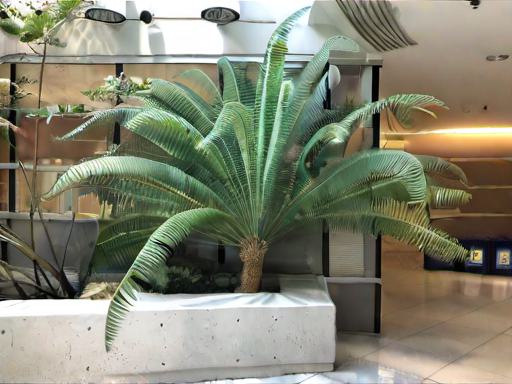} 
       }
         &
         \boxllff{
         \includegraphics[width=\magicwidth]{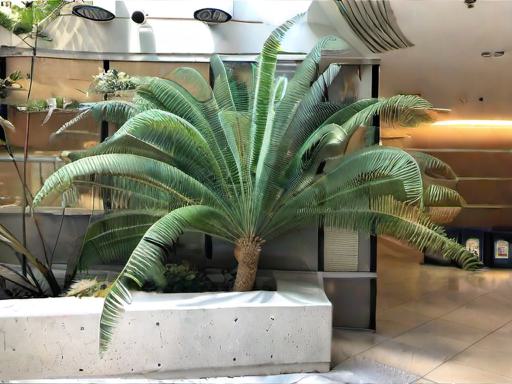} 
       }
         &
         \boxllff{
         \includegraphics[width=\magicwidth]{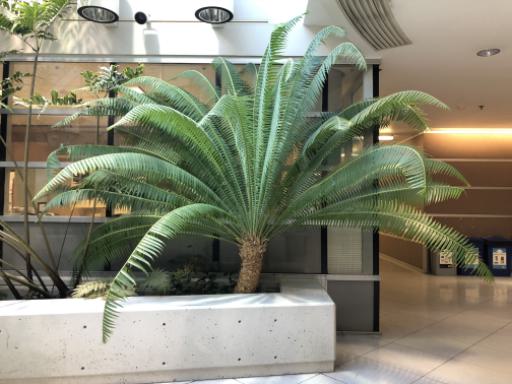}
       }
         \\

         & \multirow{1}{*}[+5em]{\rotatebox{90}{\textbf{DTU}~\cite{jensen2014large}}}
         &
         {
           \renewcommand{\arraystretch}{0.35}
           \begin{tabular}[b]{@{}c@{}}%
             \includegraphics[width=\magicthumbwidth]{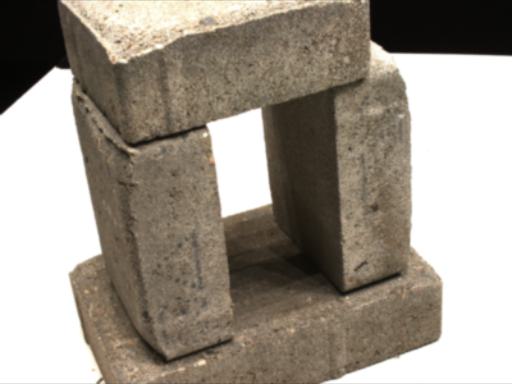} 
             \\
             \includegraphics[width=\magicthumbwidth]{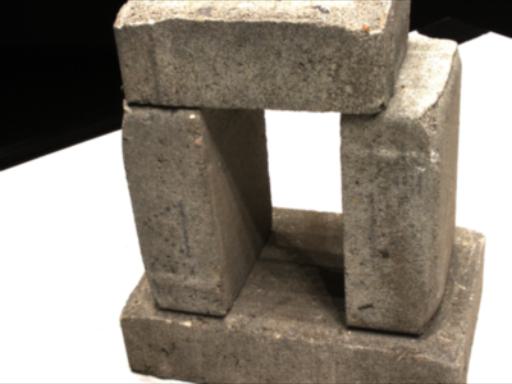} 
             \\
             \includegraphics[width=\magicthumbwidth]{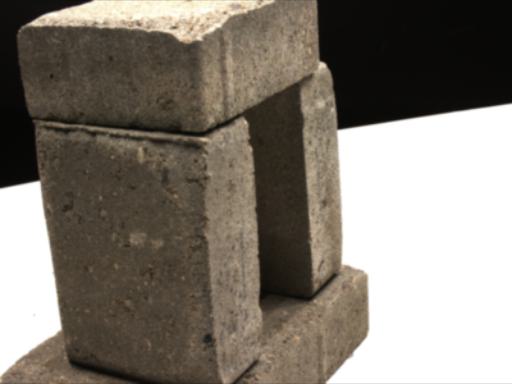}
             \\
           \end{tabular}
         }
         &
         \boxdtu{
         \includegraphics[width=\magicwidth]{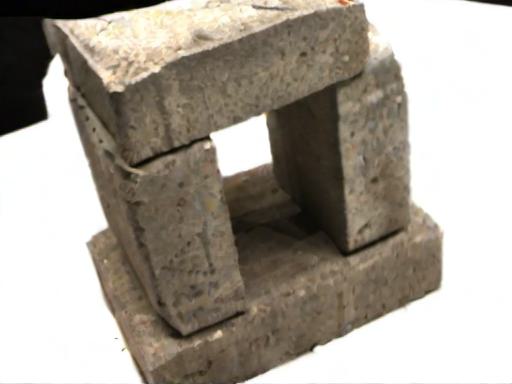} 
       }
         &
         \boxdtu{
         \includegraphics[width=\magicwidth]{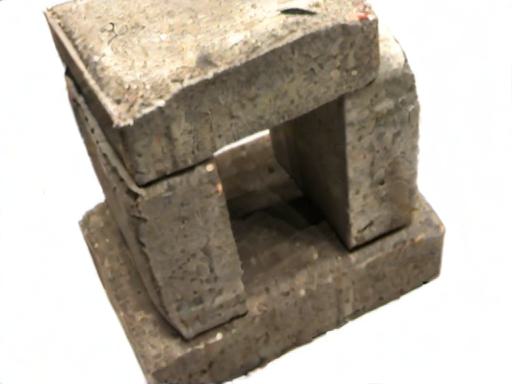} }
         &
         \boxdtu{
         \includegraphics[width=\magicwidth]{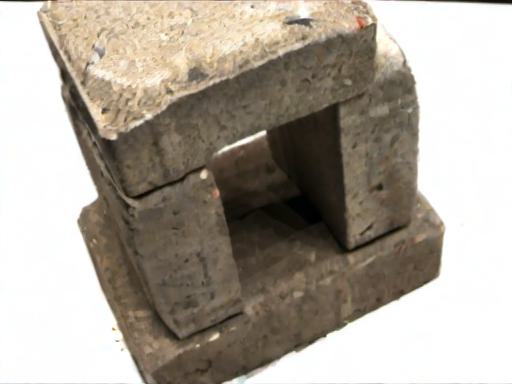} 
       }
         &
         \boxdtu{
         \includegraphics[width=\magicwidth]{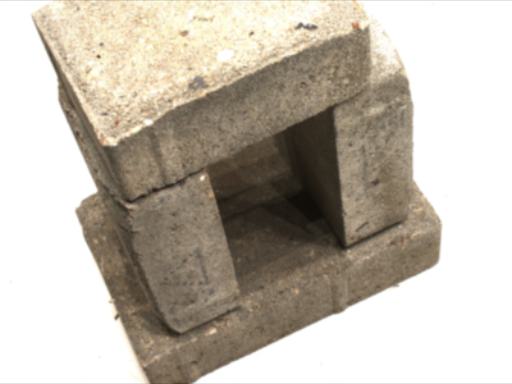}
       }
         \\
         & \multirow{1}{*}[+5.9em]{\rotatebox{90}{\textbf{Mip-NeRF}~\cite{barron22mip-nerf}}}
         &
         {
           \renewcommand{\arraystretch}{0.35}
           \begin{tabular}[b]{@{}c@{}}%
             \includegraphics[width=\magicthumbwidth]{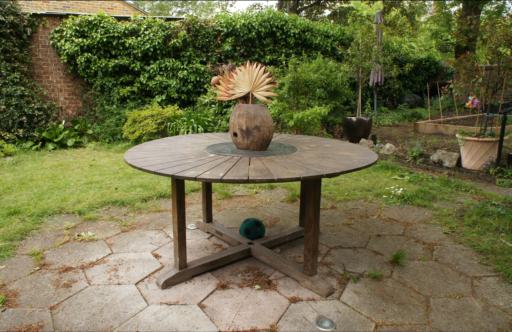} 
             \\
             \includegraphics[width=\magicthumbwidth]{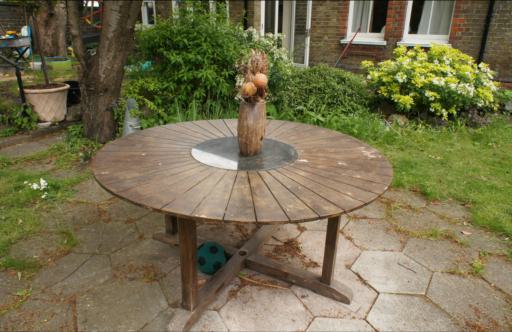} 
             \\
             \includegraphics[width=\magicthumbwidth]{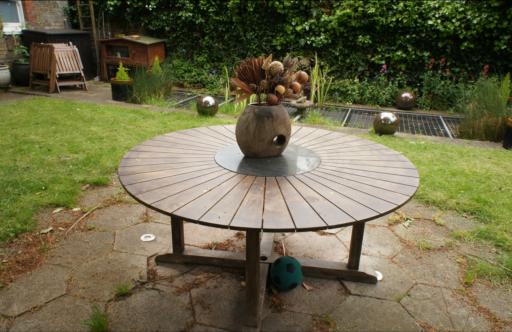}
             \\
           \end{tabular}
         }
         &
         \boxmip{
         \includegraphics[width=\magicwidth]{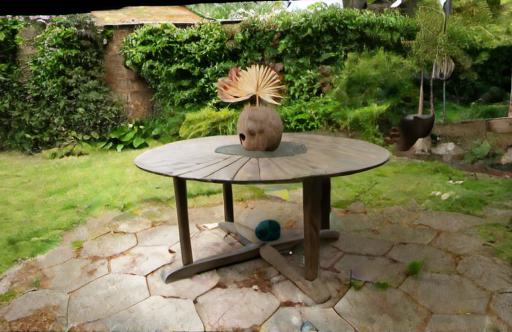} 
       }
         &
         \boxmip{
         \includegraphics[width=\magicwidth]{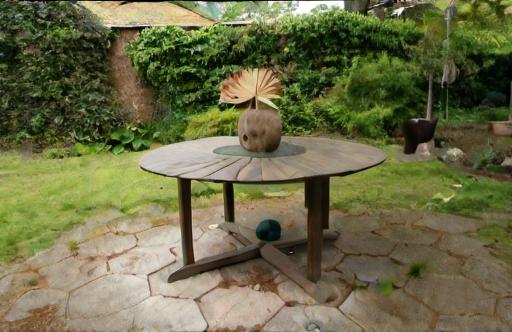}
       }
         &
          \boxmip{
         \includegraphics[width=\magicwidth]{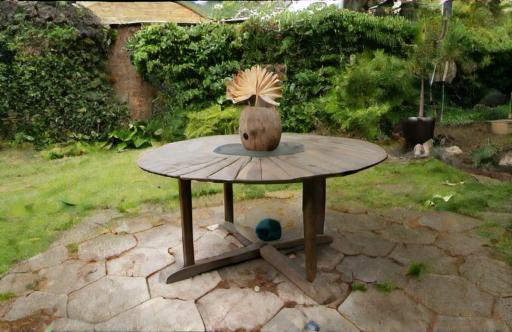} 
       }
         &
         \boxmip{
         \includegraphics[width=\magicwidth]{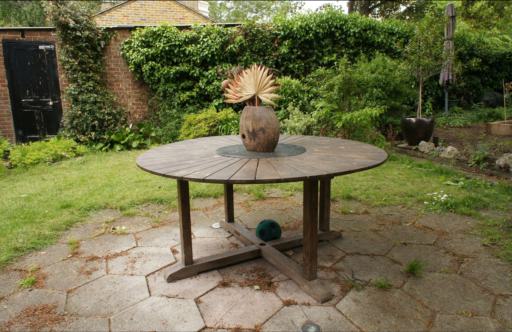}
       }
         \\

       \end{tabular}
       \caption{\textbf{Novel-view synthesis comparison.} We compare results of CAT3D-like~\cite{gao24cat3d:} models 
         trained on different datasets (MVImgNet, CO3D, \ucothreed)
         and evaluated on standard NVS datasets (top-to-bottom:
         RealEstate10K~\cite{zhou2018stereo}, 
           LLFF~\cite{mildenhall2019llff}, DTU~\cite{jensen2014large}, Mip-NeRF 
       360~\cite{barron22mip-nerf}).}%
    \label{fig:nvs_cat3d}
\end{figure*}

Sparse and even dense SfM point clouds provide an accurate but still quite sparse 3D reconstruction of the scene's surface.
To further densify it, \ucothreed provides a 3D Gaussian Splat reconstruction~\cite{kerbl233d-gaussian} for each scene, fitted using gsplat~\cite{ye2024gsplat}.

\paragraph{Scene captioning.}

\ucothreed also provides textual captions for all scenes, useful for generative modelling.
Motivated by Cap3D~\cite{luo2023scalable}, we first caption each view separately using a vision-language model, and then summarise these into a single scene caption using LLAMA3 \cite{dubey24the-llama}.
\begin{table}[t]
  \centering\small%
  \setlength{\tabcolsep}{0.05em}%
  \resizebox{\linewidth}{!}{%
  \begin{tabular}{@{}lcccccccc@{}}
    \toprule
    & \multicolumn{8}{c}{%
  \small$\underleftrightarrow{\hspace{10pt}\text{\small
  Easier}\hspace{50pt}\text{\small \textit{Dataset Difficulty}}\hspace{50pt}\text{\small 
  Harder}\hspace{10pt}}$} \\
    &
    \multicolumn{2}{c}{\textbf{Re10K}~\cite{zhou2018stereo}} & 
    \multicolumn{2}{c}{\textbf{LLFF}~\cite{mildenhall2019llff}} &
    \multicolumn{2}{c}{\textbf{DTU}~\cite{jensen2014large}} & 
    \multicolumn{2}{c}{\textbf{Mip-NeRF}~\cite{barron22mip-nerf}} \\
    Train dataset &
    LPIPS$\downarrow$ & PSNR$\uparrow$ &
    LPIPS$\downarrow$ & PSNR$\uparrow$ &
    LPIPS$\downarrow$ & PSNR$\uparrow$ &
    LPIPS$\downarrow$ & PSNR$\uparrow$ 
    \\
    \midrule
    MVImgNet~\cite{yu2023mvimgnet} &
    0.310 & 18.77 & 0.426 & 14.38 & 0.377 & 12.79 & 0.605 & 12.39
    \\
    CO3Dv2~\cite{reizenstein21common} &
    0.281 &\bf 20.02 &\bf 0.418 & 14.95 & 0.329 & 16.42 & 0.532 & 14.19
    \\
    \ucothreed (ours) &
    \bf 0.278 & 19.77 &\bf 0.418 &\bf 15.16 &\bf 0.315 &\bf 16.97 &\bf 0.528 & \bf 14.37
    \\
    \bottomrule
\end{tabular}
}
\caption{\textbf{Novel-view synthesis benchmark}. We evaluate CAT3D-like~\cite{gao24cat3d:} models trained on 
MVImgNet, CO3Dv2 or \ucothreed. We report NVS performances on RealEstate10K~\cite{zhou2018stereo}, 
LLFF~\cite{mildenhall2019llff}, DTU~\cite{jensen2014large} and Mip-NeRF 360~\cite{barron22mip-nerf}.
\label{tab:cat3d}}
\end{table}


\section{Applications}

In this section, we demonstrate \ucothreed's merit on three popular 3D learning tasks:
feedforward sparse-view 3D reconstruction (\cref{sec:feedforward_object_reconstruction}),
new-view synthesis using diffusion (\cref{sec:multiview_diffusion}), and 
text-to-3D (\cref{sec:text-to-3d}).

\subsection{Few-view 3D Object Reconstruction}%
\label{sec:feedforward_object_reconstruction}

Traditionally, multi-view 3D-annotated datasets such as CO3D or MVImgNet have been used to supervise few-view 3D reconstructors.
In this section, we train LightplaneLRM~\cite{cao2024lightplane}, an evolution of the seminal LRM~\cite{hong24lrm:}, and show that doing so on \ucothreed leads to better performance than training on alternative datasets.

LRM is a large transformer~\cite{vaswani17attention} that accepts few input images of an object and predicts a 3D representation of the latter.
The transformer, conditioned on the tokens of the observed images via cross attention, converts a set of learnable input tokens to a 3D representation.
The 3D representation is a triplane~\cite{chan22efficient} supporting an opacity/radiance implicit shape.
LightplaneLRM improves LRM by adding so called ``splatting layers'' and a memory-efficient renderer.

During training, LightplaneLRM receives four random source frames from a training \ucothreed video sequence, and renders the predicted triplane into held-out target views.
Learning minimizes the photometric loss between the renders and the corresponding ground-truth targets.
Both source and target views are masked using the extracted segmentation masks to make sure that LightplaneLRM only reconstructs the foreground object.
Training uses the Adam optimizer and is warm-started following the original LRM training protocol by pre-training the model on a large dataset of synthetic objects similar to Objaverse~\cite{deitke23objaverse:}.

\paragraph{Baselines.}

Our main goal is to demonstrate that \ucothreed contains higher quality data than existing object-centric datasets.
As such, starting from the model pre-trained on the synthetic data, we finetune either on \ucothreed, or on two other baseline datasets, namely MVImgNet and CO3Dv2.

\paragraph{Evaluation protocol.}

We evaluate each trained model in a novel-view synthesis setting on two small-scale high-quality object-centric datasets: OmniObject3D~\cite{wu2023omniobject3d} and Stanford-ORB~\cite{kuang2023stanfordorb}.
Given four views of a held-out test scene, the model reconstructs the scene which is then rendered to unseen target views.
We report the average LPIPS~\cite{zhang18the-unreasonable} loss and Peak-signal-to-noise ratio (PSNR) between each render and the corresponding ground-truth image.
We also report the intersection-over-union (IoU) between the rendered object alpha mask and the target view segmentation mask.

\paragraph{Results.}

\Cref{tab:lrm,fig:lrm} report the quantitative and qualitative results, respectively.
The LightplaneLRM trained on \ucothreed is better that the other baseines in most metrics on both datasets.
The latter confirms that \ucothreed is currently the most reliable source of real data for training feedforward few-view 3D reconstructors.

\subsection{Novel-view synthesis using diffusion}%
\label{sec:multiview_diffusion}

We now consider application of \ucothreed to training new-view image diffusion generators.
These generators can, given one or a few views of an object and a target camera pose, output new arbitrary views as observed from the target camera, hallucinating missing details based on a statistical prior they learn.
They can thus complement and integrate the feed-forward reconstruction models of the previous section, which are deterministic and thus unable to deal with ambiguity well.
To this end, we train a diffusion model similar to the recent CAT3D~\cite{gao24cat3d:}, but reimplement it from scratch given lack of source code (see details in the supplementary).
We call this model CAT3D-like.

\paragraph{Evaluation protocol.}

As in \cref{sec:feedforward_object_reconstruction}, we compare versions of CAT3D-like trained using \ucothreed, MVImgNet, and CO3Dv2 and test them on held-out datasets.
A feature of CAT3D is the ability to reconstruct both the principal object in the images as well as the background.
We thus benchmark the method using new-view synthesis datasets that do contain background, namely
DTU~\cite{jensen2014large}, containing structured light scans of various objects,
LLFF~\cite{mildenhall2019llff}, containing scenes captured from fronto-parallel camera trajectories,
RealEstate10k~\cite{zhou2018stereo}, containing real-estate walkthroughs,
and Mip-NeRF 360 \cite{barron21mip-nerf:} with complex indoor and outdoor scenes.
For evaluation, we take three known views as input and use the model to predict a new view.
We report LPIPS and PSNR but not the IoU since CAT3D only generates new RGB views without reconstructing the 3D shape.

\begin{figure}
\centering
\includegraphics[width=\linewidth]{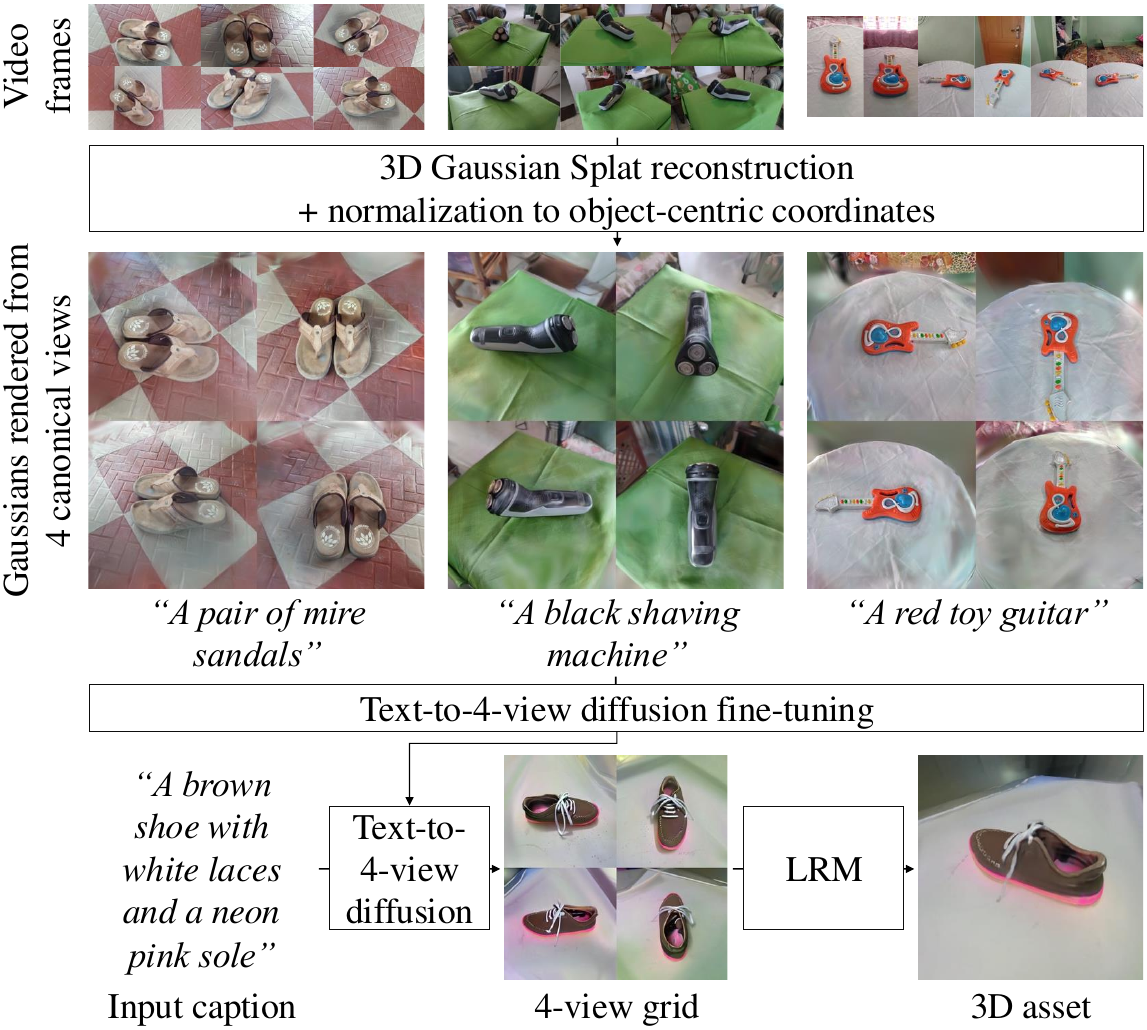}
\caption{\textbf{Supervising Instant3D with 3DGS.}
For each training scene, its 3DGS is rendered from 4 canonical views yielding a captioned image dataset for finetuning an image diffuser.
Samples from the latter are then reconstructed with LRM.}%
\label{fig:instant3d}
\end{figure}

\paragraph{Results.}

\Cref{tab:cat3d,fig:nvs_cat3d} contain the results:
training CAT3D-like on \ucothreed leads to the best performance across all four datasets.
Even when compared to MVImgNet, which is slightly larger than \ucothreed, the latter improves PSNR by 3--4 points, and reduces the LPIPS error by 5\% to 20\%.

\subsection{Photorealistic Text-to-3D}%
\label{sec:text-to-3d}

\begin{figure*}[t]%
\centering%
\includegraphics[width=\linewidth]{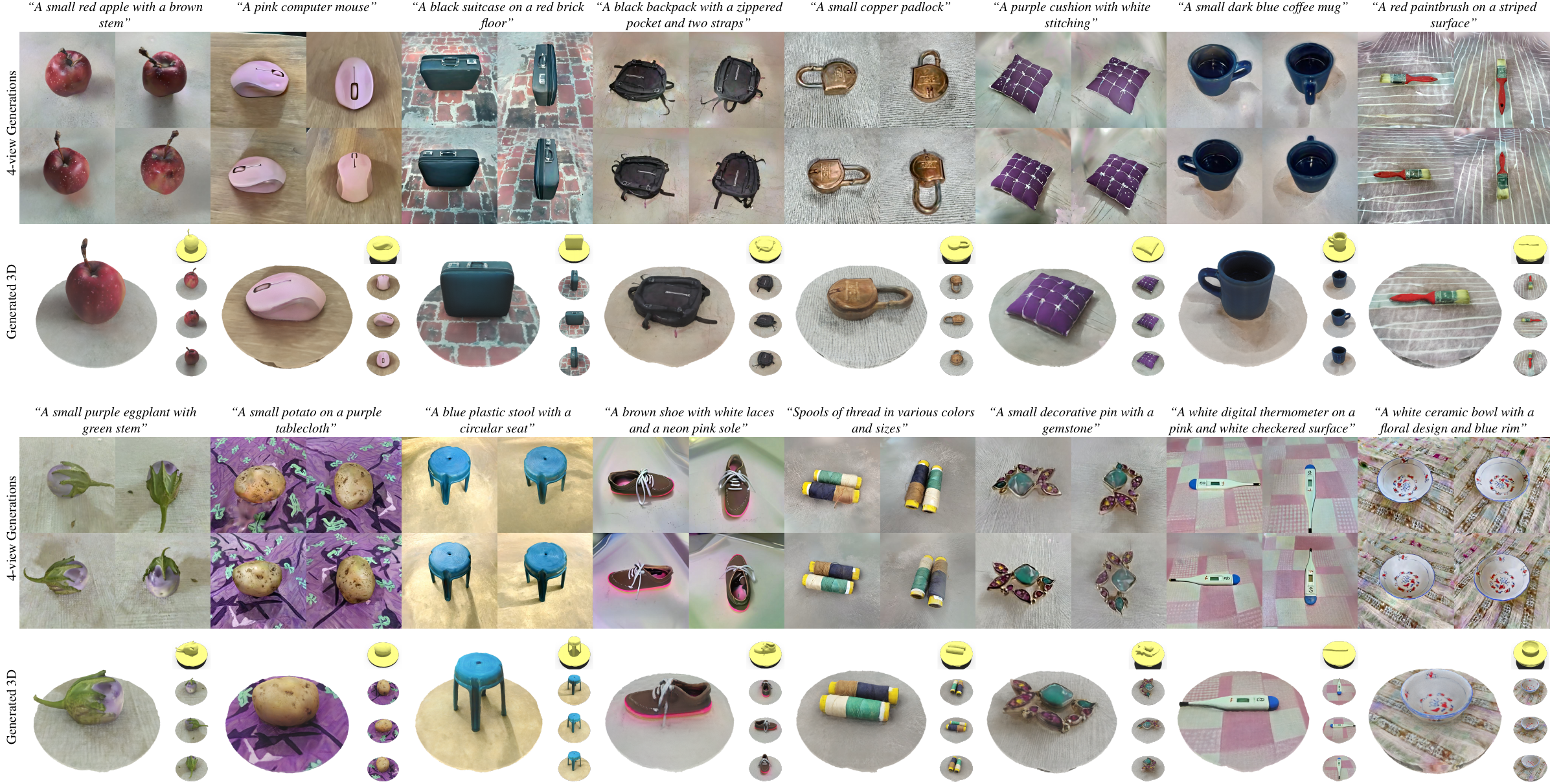}%
\caption{
\textbf{Qualitative results for text-to-3D} generation displaying the 4-view grids generated by our Instant3D-like model given the input caption, and the 3D asset obtained by reconstructing the latter.
The 4-view grid generator was trained using the canonical 4-view renders of \ucothreed's 3DGS scene reconstructions.
\label{fig:textto3d}%
}%
\end{figure*}

Next, we show that \ucothreed allows to train a photorealistic text-to-3D generators.
Methods like CAT3D and others~\cite{liu23zero-1-to-3:,shi24mvdream:,melas-kyriazi24im-3d} generate several views of the object first, and then fit a 3D model, such as a NeRF or 3DGS, via optimization.
This can work well, but it is not particularly robust or fast.
An alternative, popularized by Instant3D~\cite{li24instant3d:} and follow-ups~\cite{siddiqui24meta,xu2024grm}, is to use a feedforward reconstructor in the second step, similar to LightplaneLRM from \cref{sec:feedforward_object_reconstruction}, which is faster and more robust.
However, these models require \emph{canonical} views of the objects --- for example, Instant3D considers 4 orthogonal viewpoints, covering all `sides'.
The requirement of such training canonical views complicates training on real data, where viewpoints are arbitrary, and explains why such models are usually trained on synthetic data, limiting realism.


\paragraph{Imaging 3DGS from canonical views.}

Our new idea is to `re-shoot' the 3DGS reconstructions provided with \ucothreed from canonical viewpoints, making our data compatible with any method requiring canonical views for training (\cref{fig:instant3d}).
To do so, we render the normalized reference frames (\cref{sec:uco3d}) into four views for each object, and arrange them in a grid as a target for the text-to-4-view generator.
We double check the quality of the renders by calculating their CLIP similarity~\cite{hessel2021clipscore} to the object caption, and discard a sample if this is below 0.3.

We use this data to fine-tune a text-to-4-view image diffusion model using these 4-view image grids and the corresponding scene captions.
At inference time, given a caption describing the desired object, we use the model to sample a new 4-view grid and feed the latter, together with the corresponding cameras, to the LightplaneLRM model (\cref{sec:feedforward_object_reconstruction}) for 3D reconstruction.



\paragraph{Baselines.}

We train another 4-view generator on a dataset of synthetic assets similar to Objaverse~\cite{deitke23objaverse:} and use it with the original LightplaneLRM model~\cite{cao2024lightplane} trained on the same data and thus optimally matched to it.

\begin{table}[t]
  \centering\footnotesize%
  \begin{tabular}{@{}ccc@{}}
  \toprule
  Train dataset & \textbf{Real} - FID$\downarrow$  & \textbf{Surreal} - FID$\downarrow$ \\
  \midrule
  Synthetic & 82.8 & 42.3\\
  \ucothreed (ours)  & 63.9 & 68.9 \\
  \bottomrule
\end{tabular}
\caption{%
\textbf{Text-to-3D evaluation.} We compare Instant3D-like models trained on \ucothreed or a dataset of synthetic renders from artist-created meshes. We report FID on two sets of data corresponding to real and surreal objects, see text for details.
}%
\label{tab:textto3d}
\end{table}



\paragraph{Evaluation protocol.}

We report metrics evaluating the alignment between the distributions of the generations and the ground-truth objects.
Specifically, we report the Frechet Inception Distance (FID)~\cite{heusel17gans} between the renders of the generated 3D shapes and images of ground-truth objects.


The main purpose of this experiment is to show that, by training on the \ucothreed dataset, the 3D generations are more realistic.
We assess this using two sets of prompts:
\textbf{Surreal}, containing 100 captions of objects from the synthetic dataset,
and \textbf{Real}, containing 100 random captions form the held-out evaluation sequences of \ucothreed.
We report FID between the generated 3D shapes and the images/renders corresponding to the objects of each prompt-set.


\paragraph{Results.}

\Cref{tab:textto3d,fig:textto3d} contain the quantitative and qualitative results respectively.
The table reveals that the \ucothreed -trained generator outperforms the synthetic generator when evaluated on real prompts.
The latter verifies our hypothesis that a generator trained on \ucothreed yields more realistic samples than a model trained on synthetic data.

\section{Conclusions}%
\label{sec:conclusions}

We have introduced \ucothreed, a new object-centric 3D dataset of real-life objects.
\ucothreed strikes a balance between size and quality, ensuring the quality of the collected turntable-like videos and of the 3D annotations, while at the same time significantly expanding the scale of the data compared to CO3Dv2 and the diversity compared to CO3Dv2 and MVImgNet.
We have shown the benefits of using this dataset compared to alternative when training models for feedforward few-view 3D reconstruction, multi-view generation, and text-to-3D generation.
Equipped with 3D cameras, point cloud, masks, textual captions and 3DGS reconstructions of objects, \ucothreed is a ready-to-use resource for training large generative models and for exploring 3D deep learning.
\appendix
\section{Appendix}

\subsection{CAT3D-like model details}
This section provides more details for the CAT3D-like model used in~\cref{sec:multiview_diffusion}. 
CAT3D~\cite{gao24cat3d:}~is a diffusion model which takes as input a set of $N_\textrm{tgt}$ target cameras and a set of $N_\textrm{src}$ source views with cameras, and aims at generating the views associated to the target cameras.
Since the code is not available, we follow the implementation details of~\cite{gao24cat3d:} to reproduce a model with similar capabilities.
Specifically, starting from a pretrained text-to-image latent diffusion model similar to~\cite{rombach22high-resolution}, we first modify all 2D self-attention layers in the decoder part of the denoising UNet such that 2D self-attention is performed across all the views in the batch. First proposed by~\cite{shi24mvdream:}, this cross-view attention allows each image token to attend to tokens of all views in the batch, thus improving multi-view consistency. 
Then, we modify the architecture with zero-initialized channel expansion such that it can take as input the latent features concatenated with mask maps indicating source views and camera maps in the form of Plücker rays.
Different from~\cite{gao24cat3d:}, we use the v-prediction / v-loss parametrization~\cite{salimans22progressive} and the zero terminal SNR noise scheduling recommended by~\cite{lin23common} as we found it to work better than the original CAT3D recipe.
For all experiments, we use $N_\textrm{src} = 3$, $N_\textrm{tgt} = 5$ and train for 100k iterations using Adam~\cite{kingma14adam:} optimizer with a constant learning rate of $1e^{-5}$ and a global batch size of 64.

For evaluation, we follow standard practices~\cite{wu2023reconfusion,gao24cat3d:} and report results on common out-of-distribution NVS datasets (RealEstate10K, LLFF, DTU, Mip-NeRF 360) using the same test splits. We evaluate novel-view synthesis in the 3 view input setting and report LPIPS and PSNR metrics.

\subsection{Text-to-3D model details}
The text-to-image stage of the Instant3D-like model from \cref{sec:text-to-3d} is based on an internal text-to-image model architecturally similar to Emu~\cite{dai23emu:}. Starting from a model pre-trained on a dataset of image-caption pairs, we fine-tune the model on 4-view canonical-render grids of \ucothreed objects.
In all our experiments we use the Adam optimizer ~\cite{kingma15adam:} with a batch size of 160 and a constant learning rate of $1e^{-5}$.
We distribute the training across 32 NVIDIA A100 gpus for a total of 20k steps.
During inference, we use a Diffusion Probabilistic Model (DPM) Sampler~\cite{lu22dpm-solver:} and denoise over 60 steps.
The 4-view-to-3D stage of the Instant3D-like model is based on LightplaneLRM~\cite{cao2024lightplane}.
For the LightplaneLRM model which reconstructs both the central object and the scene background, we use the coordinate contraction of MERF \cite{reiser23merf:} which non-linearly maps the distant parts of the scene so they always fall into the [-1,1] bounding cube of the utilizied triplane representation.
The rest of the training procedure follows the LightplaneLRM protocol described in \cref{sec:feedforward_object_reconstruction}.

We train three different versions of both the Instant3D-like model and LightplaneLRM, each version corresponding to a different dataset. Specifically, we train on a dataset of synthetic assets similar to Objaverse~\cite{deitke23objaverse:}, and on two versions of \ucothreed, one that contains background and one where the background information is masked. For evaluation, we report the FID metric~\cite{heusel17gans} for the models trained on datasets without background information (\cref{tab:textto3d}). The evaluation sets corresponding to the \textbf{Surreal} and \textbf{Real} prompts are created by randomly selecting 50 image frames for every scene / prompt pair. We center the objects of the \ucothreed images using the per-frame mask information for consistent evaluation across all datasets. The generated 3D objects of the text-to-3D model are rendered from sampled cameras drawn from the camera distribution of each individual evaluation set. The qualitative results presented in \cref{fig:textto3d} are extracted using the model variants trained with background information. 


\subsection{Rigid scene alignment}
In \cref{sec:uco3d}, we described a procedure that estimates a rigid transform for each object to align it to dataset-wide object-centric reference.
Here, we provide additional details.

We start with finding the gravity axis by making sure the roll of the cameras is close to 0, following~\cite{szymanowicz23viewset}.
Then, we translate and scale the densified point cloud ((b) in \cref{fig:uco3d_reconstructions}) so that the median locations along the horizontal axes are 0, and so that the STD of its points' coordinates is 1.
Then, we normalize the 2D rotation in the horizontal plane by aligning the principal components, and, finally, shift the object vertically to make the ground plane's elevation zero.
As shown in the experiments, this normalisation allows to render each object from 4 canonical viewpoints defined in the object-centric reference, which eventually enables the Instant3D-like text-to-3D model training.

{%
\small%
\bibliographystyle{ieeenat_fullname}%
\bibliography{main,vedaldi_general,vedaldi_specific,metallrm}%
}%
\end{document}